# TSBPO: Tree-Structured Branching for Policy Optimization in Interactive Language Agents

**Rui Yu, Chong Mu, Jinchuan Zhang***
University of Electronic Science and Technology of China
Rui Yu: 202522081031@std.uestc.edu.cn | Jinchuan Zhang: jc.zhang@uestc.edu.cn
*Corresponding author: Jinchuan Zhang

## Abstract

Training interactive language agents from terminal outcomes poses three coupled problems: which task and state continuations to sample, how to assign delayed feedback to decisions, and how to aggregate token losses for multi-token actions. We present TSBPO, a tree-structured sampling-to-update paradigm that addresses these problems within one training loop. A likelihood-fitted Root/Branch allocator uses retained task and state evidence to target continuations likely to produce contrasting terminal outcomes under fixed rollout budgets. On the resulting replayed physical trees, trajectory advantages compare immediate sibling segments, while observation-group step advantages compare outcomes at matched task/observation states. The actor then sums clipped token surrogates over each complete response and normalizes across active turns, avoiding response-length averaging in the policy-gradient term. With a Qwen2.5-1.5B actor, fixed checkpoints obtain 97.79% seen and 96.64% unseen stochastic success on ALFWorld and an 81.07% per-episode success rate over six stochastic episodes for each of the 500 WebShop test goals. A WebShop step-term ablation yields 75.93% at the same update-100 evaluation; an ALFWorld aggregation comparison shows higher observed command validity with complete-turn weighting. These results support the evaluated training configuration on two interactive benchmarks.

## 1. Introduction

Interactive language agents act through generated commands and receive terminal task feedback after partially observed, variable-length interactions. ALFWorld and WebShop make completion verifiable [1, 12], but a terminal label alone does not identify where additional rollouts should be spent, which turns deserve credit, or how a multi-token response should enter an actor update. The prompt is a bounded account of observations and earlier commands, so replaying a physical occurrence differs from merely matching its current text [25, 26].

TSBPO connects these decisions in one training process. Supervised protocol alignment initializes executable behavior; a Root/Branch curriculum acquires outcome alternatives; realized physical siblings and exact observation groups supply separate credit terms; complete-turn aggregation transfers that credit to token updates. Both evaluated runs use the likelihood-fitted Root/Branch allocator and local state-group contrast objective, with benchmark-specific rollout budgets. Their task results measure the integrated training process rather than the isolated effect of a single component.

We evaluate fixed Qwen2.5-1.5B checkpoints on complete ALFWorld validation sets and all 500 WebShop test goals. Training diagnostics characterize acquired comparisons and prediction reliability; configuration comparisons describe specific optimization choices. The derivations, full protocols, and additional studies appear after the references.

## 2. Related Work

### 2.1 Interactive policies and terminal feedback

Interactive language agents choose natural-language responses that also act on an environment. ReAct interleaves reasoning and actions with observations [13], while Reflexion uses textual feedback to revise subsequent behavior [14]. Both illustrate the importance of a closed-loop interaction history. Policy training adds a different problem: a completed trajectory provides an outcome, but that outcome does not identify which earlier commands or sampled continuations were most useful. ArCHer studies hierarchical multi-turn reinforcement learning [20], long-horizon interactive RL addresses sustained environment interaction [21], and RAGEN examines multi-turn training dynamics [11]. In ALFWorld and WebShop, task completion is measured after a sequence of agent actions [1, 12]; the terminal result therefore summarizes a variable-length interaction rather than an individual response.

Partial observability makes the agent's history, rather than the latest observation alone, relevant to action selection [25]. Temporal abstraction separates decisions made at the environment timescale from the lower-level operations used to realize them [26]. For a language-model actor, one environment action may contain many autoregressive tokens, whereas the environment transition occurs after the response is submitted. This separation creates three distinct training decisions: sampling determines which future outcomes are observed, credit assignment compares realized actions and continuations, and policy-gradient aggregation determines how those comparisons weight token updates. TSBPO retains a physical replay structure for the first two decisions and treats the complete response as the unit of actor optimization.

## 2.2 Curriculum and rollout allocation

Curriculum learning adapts the experience distribution to the learner [29, 30]. DPS predicts fully unsolved, mixed, and fully solved prompt groups through a three-state task-dynamics model [9]. Sparse revisits motivate retaining direct task evidence and borrowing capped evidence from similar instructions. Branch acquisition uses likelihood-fitted state evidence at the original remaining deadline to allocate continuations toward terminal-label contrasts in factual-homogeneous observation groups.

Hindsight Credit Assignment relates outcome-conditioned occurrence probabilities and forward outcome probabilities through Bayes' rule [23]. This connection informs the use of completed outcomes for continuation prediction. In TSBPO, predictive evidence determines sampling before the actor's advantages exist; fresh observed outcomes determine credit afterward. This separation keeps the acquisition model from becoming an additional reward or advantage multiplier.

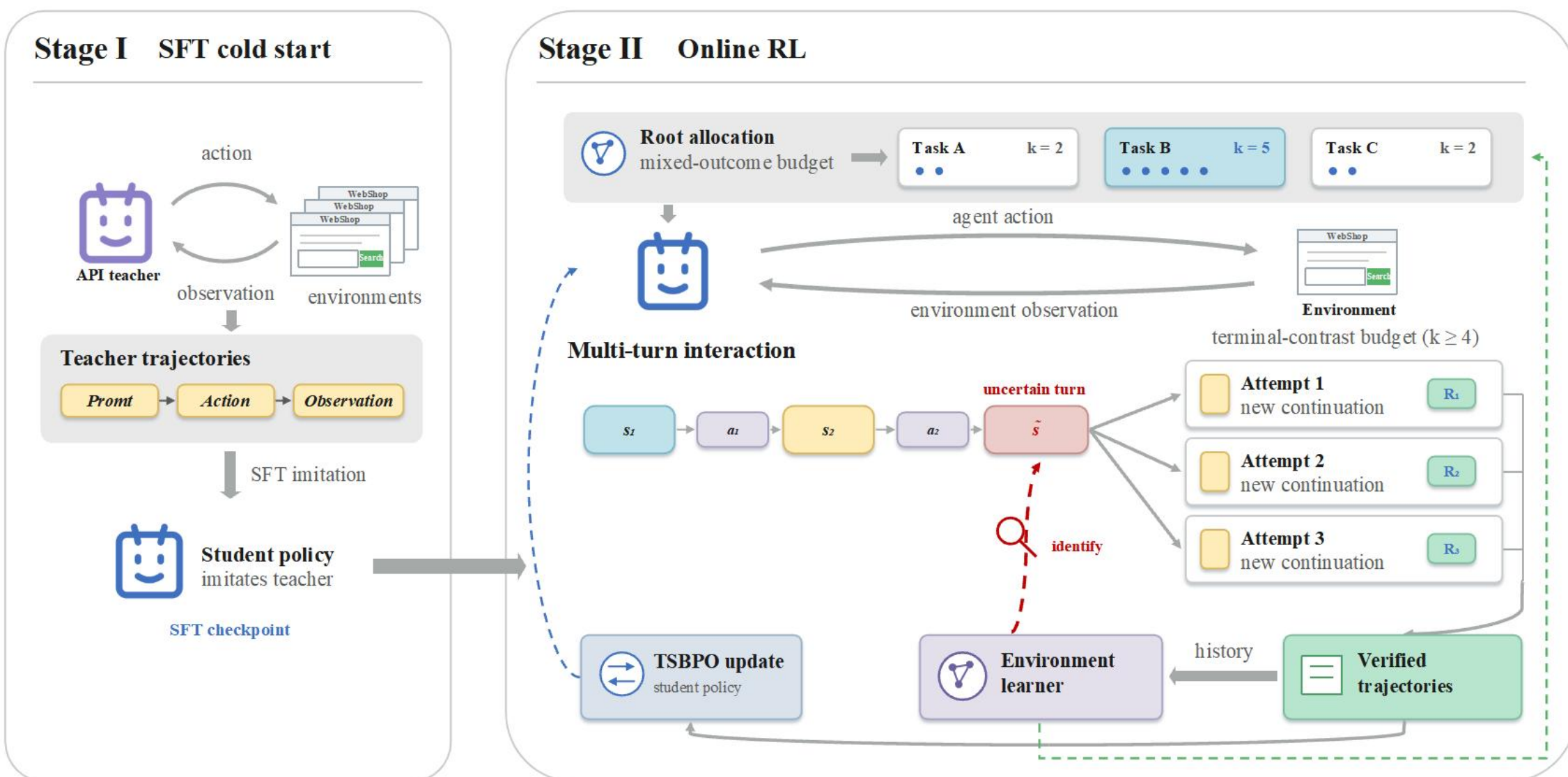

Figure 1. TSBPO training paradigm. Successful teacher interactions initialize executable behavior; online RL connects Root/Branch allocation, physical-tree credit, and complete-turn actor updates. Rollout icons are schematic.

## 2.3 Local credit and relative optimization

GiGPO combines episode-relative and observation-group step advantages [4, Section 4]. HGPO refines equal-current-observation groups by shared historical context and combines their advantages with depth-dependent weights; K denotes the maximum historical context depth [33]. SALT refines outcome-derived advantages on a trajectory graph [31], while GraphGPO credits transitions using estimated progress toward a goal [32]. These methods construct comparisons from existing rollouts. TSBPO additionally acquires alternatives by replaying physical occurrences; its trajectory comparison uses immediate siblings sharing the realized history, alongside the retained observation-group step term.

DPEPO [34] trains an agent to issue actions across multiple environments during inference, using a trajectory-level success reward together with action- and state-transition-diversity rewards. TSBPO trains a sequential interaction policy: replayable physical occurrences organize training continuations, and observed terminal outcomes determine local sibling credit. The methods therefore use multiple environment trajectories at different stages and optimize different interaction interfaces.

TreePO aggregates relative rewards across nested ancestor subgroups [5], whereas our segment credit ends at the next physical split and restarts there. AT²PO couples entropy-guided expansion of turn-level rollout trees with descendant-based value propagation and turn-level importance weighting [6]. AT²PO and TSBPO both acquire tree-structured continuations, but our branch selection estimates deadline-conditioned terminal-label contrast; trajectory credit compares immediate physical siblings alongside a separate observation-group step term. BiPACE refines groups through behavioral state neighborhoods and action-conditioned peers [8]. RUDDER learns return redistribution for delayed feedback [22]; here, terminal leaves supply segment outcomes directly.

REINFORCE and baseline analyses establish how sampled outcomes enter score gradients [27, 28]. Leave-one-out estimators distinguish an independent baseline from a sample mean containing the scored action [18, 19]. PPO and value-based estimators offer established update and credit mechanisms [2, 16, 17]. Dr. GRPO identifies response-length and reward-normalization factors as changes in optimization weighting [7], and DAPO revises token aggregation [10]. We adopt a shared active-turn denominator while summing token surrogates within each response. The resulting local score identity explains the interface between tree credit and the language policy; Appendix B treats its conditions and adaptive-sampling limits.

## 3. Interaction Setting and Training Paradigm

### 3.1 Complete turns and observed state groups

Let μ be the evaluation distribution over tasks. For task x, the actor receives context $H_i$, generates response $y_i$ containing reasoning and an environment command, observes the resulting transition, and continues until termination or the action limit. A trajectory τ includes generated responses and environment transitions; the verifier supplies $R(\tau)\in\{0,10\}$. The task objective is

$$J_\mu(\theta)=E_{x\sim\mu}\,E_{\tau\sim P_{\pi_\theta}(\cdot|x)}[R(\tau)]. \quad (1)$$

The curriculum changes the training task distribution and selects physical turns for an empirical actor surrogate. Terminal outcomes define task success; local invalid-action costs and fixed-reference regularization serve separate training roles. Adaptive sampling makes this surrogate distinct from the evaluation objective.

At the language-policy level, each response factorizes autoregressively:

$$\log\pi_\theta(y_i|H_i)=\sum_{j=1}^{L_i}\log\pi_\theta(y_{i,j}|H_i,y_{i,<j}). \quad (2)$$

The prompt $H_i$ contains the current observation and bounded interaction history; generated EOS tokens are included, while generation-limit termination adds no EOS. Exact task/observation keys define statistical state groups. Replay additionally retains a physical occurrence's environment snapshot and realized history.

### 3.2 Supervised protocol alignment

Successful trajectories collected with GPT-5.6-sol provide environment-specific user–assistant pairs. Starting directly from Qwen2.5-1.5B-Instruct, two epochs of causal cross-entropy on assistant tokens align the initial policy with the interaction protocol:

$$L_{SFT}(\theta)=-E_{(H,y)\sim D_T}[\textstyle\sum_j\log\pi_\theta(y_j|H,y_{<j})]. \quad (3)$$

The supervised model initializes both the trainable actor and its fixed RL reference. The teacher supplies no online RL feedback; the environment supplies terminal labels. Figure 1 summarizes the two-stage process, and Appendix A describes collection, prompts, and supervised settings.

### 3.3 Frozen behavior policy and the physical rollout tree

A replayed factual prefix is stored once. Each physical split retains one factual continuation and adds new sibling suffixes. A child segment runs until the next split or a terminal leaf, so its turns receive one local comparison. A deeper split starts another comparison rather than accumulating ancestor advantages. Figure 2 also distinguishes optional actor-side coefficient weighting from the raw trajectory and observation-group credit terms.

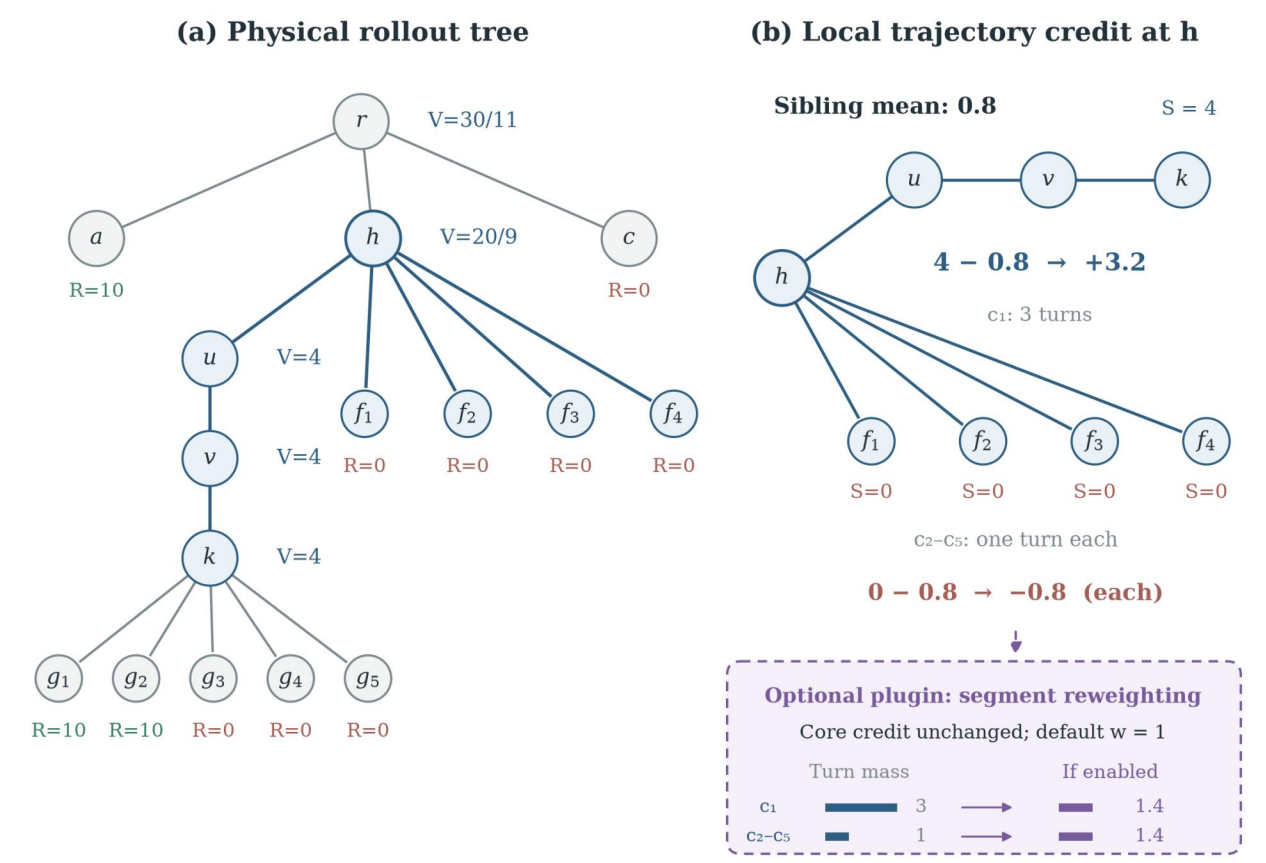

Figure 2. Physical rollout tree and local trajectory credit. Terminal leaves back up as descendant-leaf means; immediate sibling scores are centered at each split. At h, the child scores 4 against four zero-score siblings, yielding +3.2 and -0.8. Optional weighting changes actor coefficients only.

Each round freezes a behavior policy $\pi_{old}$ for factual trajectories and new suffixes. Replay reconstructs a visited physical occurrence; shared prefixes are stored once, and replay itself creates no new training row. Every newly generated physical turn supplies one active actor row. After sampling finishes, observed returns determine detached advantages, which enter the complete-turn actor update. Prediction chooses what to sample and never scales the resulting credit.

## 4. Dynamic Curriculum for Acquiring Outcome Comparisons

Root assigns factual episodes to tasks; Branch assigns new suffixes at replayed physical occurrences before advantages are computed. Both seek terminal-label contrasts under fixed rollout budgets. A mean success rate near one half can conceal nearly deterministic tasks, so separate likelihood-fitted Root and Branch models integrate uncertainty in task or state success probabilities. They weigh aged direct evidence and bounded semantic evidence; Branch also conditions on current factual labels and the original remaining deadline. Both evaluated runs use this path. Likelihood fitting alone does not guarantee calibration.

### 4.1 Root selection and fitted factual quotas

For task i, let $h_i^s,h_i^f$ be age-decayed direct success/failure counts (factor 0.95 per round), and $m_i^s,m_i^f$ be capped counts retrieved from similar task instructions. Each parameter component $\theta=(a_\theta, b_\theta, \eta_\theta, \rho_\theta)$ defines $p_i|\theta,D\sim \mathrm{Beta}(A(i,\theta),B(i,\theta))$, where $A(i,\theta)=a_\theta+\eta_\theta h_i^s+\rho_\theta m_i^s$ and $B(i,\theta)=b_\theta+\eta_\theta h_i^f+\rho_\theta m_i^f$. The weights $\eta_\theta$ and $\rho_\theta$ lie in [0,1], so historical and semantic evidence can be discounted separately. Root decisions have no current factual counts.

The Root mixture weights $w_i(\theta)$ fit complete physical Root-group counts by beta-binomial predictive likelihood. Each past group receives age weight $0.95^{age}$ divided by its task's group count that round; fitting excludes the queried task. Unidentified semantic or historical reliability uses $\rho_\theta=0$ or $\eta_\theta=1$, respectively; scant population evidence uses Beta(1/2,1/2). Retrieved counts inform the query without entering direct outcome memory.

A candidate pool combines fresh task instances with a bounded set of previously evidenced instances, allowing useful tasks to reappear despite sparse natural revisits; historical candidates occupy at most half the pool. Let $g_i=U_{root}(i,n_{ref})$, where $n_{ref}$ is the factual group size. Tasks are drawn without replacement using weights proportional to

$$P(i)=\varepsilon/M+(1-\varepsilon)\exp(g_i/T_c)/\textstyle\sum_u \exp(g_u/T_c). \quad (4)$$

We use $\varepsilon=0.30$ and $T_c=0.25$. For $n\geq1$ outcomes that are conditionally independent given a task success probability, the event probability within a Beta component is

$$u(n;A,B)=1-B(A+n,B)/B(A,B)-B(A,B+n)/B(A,B).$$

Averaging this event probability over the fitted task-specific components gives the Root acquisition curve:

$$U_{root}(i,n)=\textstyle\sum_\theta w_i(\theta)u(n;A(i,\theta),B(i,\theta)). \quad (5)$$

Here B is the beta function and $U_{root}(i,0)=0$. Integrating each component before averaging avoids treating an uncertain mean success rate as a known task probability. Each selected task receives at least two factual rollouts; the remaining Root budget follows marginal gains in (5). The same fitted mixture determines the curriculum score and the quota curve. Appendix B.1 derives the component event probability, and Appendix C.4 specifies semantic retrieval.

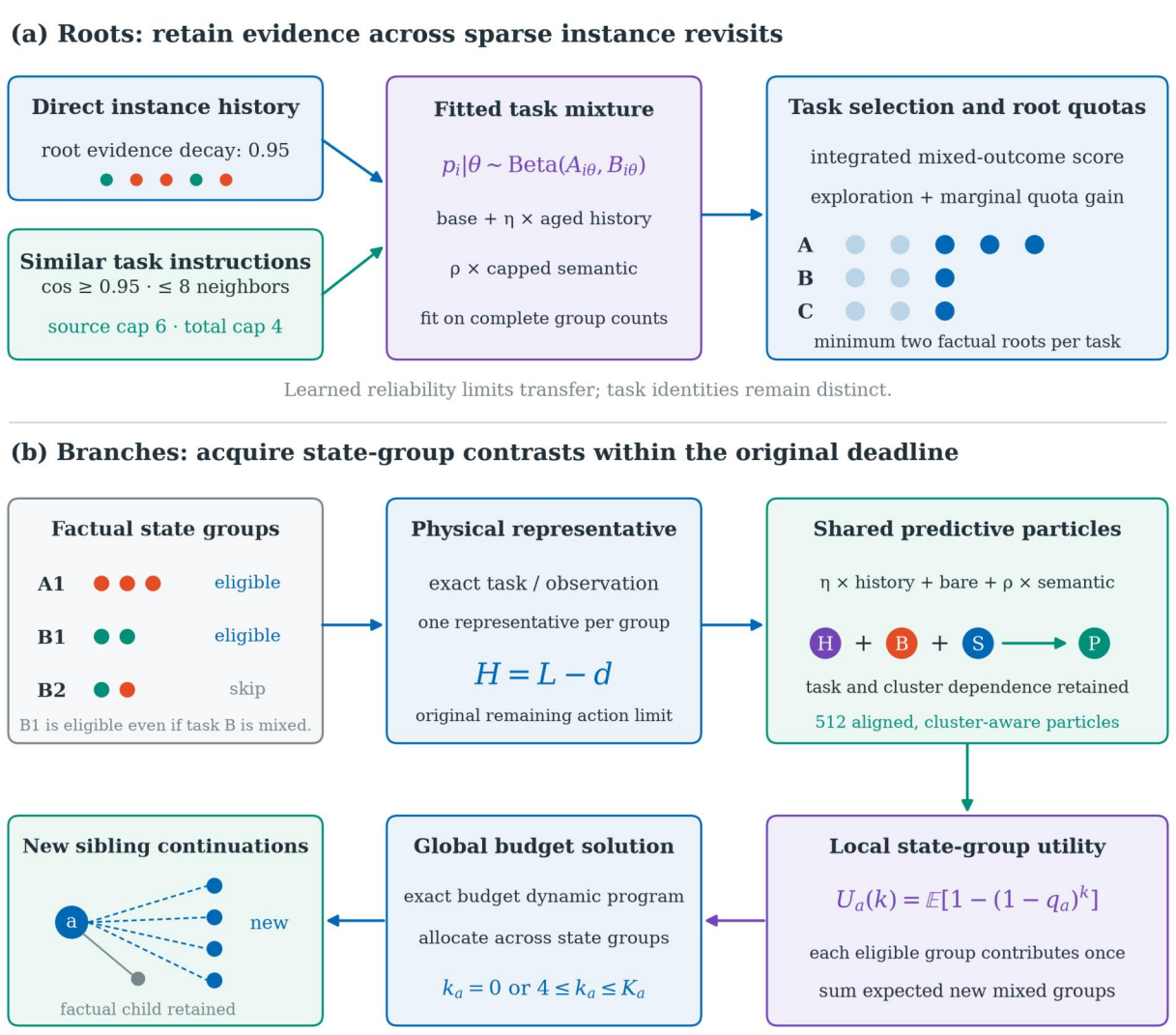

Figure 3. Likelihood-fitted Root/Branch allocator used in ALFWorld and WebShop. Root assigns factual quotas from decayed direct and capped semantic evidence. Branch scores eligible exact observation groups under original deadlines and solves a global suffix budget with at least four new continuations per opened group.

## 4.2 Deadline-conditioned Branch prediction

Completed factual trajectories form exact task/observation groups. A group is eligible when its factual labels are homogeneous, a physical occurrence is replayable, and its predicted opposite-outcome probability is positive; the task may already be mixed overall. The earliest eligible occurrence is its representative. At depth d, suffixes retain the original deadline H=L−d.

For a state/depth pool a, let $h_a^s, h_a^f$ denote historical direct counts after age decay (factor 0.90 per round), $c_a^s, c_a^f$ the completed current bare counts, and $m_a^s$, $m_a^f$ capped semantic-neighbor counts gated by task goal, state similarity, and deadline. Each physical trajectory contributes at most once to a pool. For concise notation, the component-level success/failure mass summaries are

$$\begin{aligned} A(a,\theta)&=a_\theta+\eta_\theta h_a^s+c_a^s+\rho_\theta m_a^s,\\ B(a,\theta)&=b_\theta+\eta_\theta h_a^f+c_a^f+\rho_\theta m_a^f. \end{aligned} \quad (6)$$

Branch fits a separate mixture to complete historical Branch groups by an age- and task-balanced beta-binomial likelihood, holding out the queried task. When Branch history is sparse, factual Root groups from other tasks initialize only the base success-rate distribution; Root reliability coefficients are not transferred. Current bare counts enter (6) once; learned $\eta_\theta$ and $\rho_\theta$ scale historical and borrowed evidence only. Joint acquisition uses correlated samples rather than independent Beta marginals.

## 4.3 State-group contrast utility and branch quotas

Let $A_{hom}$ contain the factual-homogeneous candidate state groups. For group a, $q_a$ is the modeled probability that one new suffix ends with the label opposite to its observed factual label: $q_a=p_a$ after factual failure and $q_a=1-p_a$ after factual success. Let $k_a$ be the number of new suffixes assigned to a. The expected probability of creating a mixed terminal-label group is

$$U_a(k_a) \; = \; E[1-(1-q_a)^{k_a} \mid D]. \quad (7)$$

Given evidence D, $U_a(k_a)$ averages the event probability over 512 aligned posterior particles. Each eligible observation group counts once, including distinct groups within one task. Suffix labels are conditionally independent within a particle, while shared task and trajectory-cluster draws retain modeled dependence. Groups already mixed at their own observation are ineligible.

For a block of m additional suffixes at group a after its current quota $k_a$, the local gain is

$$\Delta_{a,m}(k_a) \; = \; E[(1-q_a)^{k_a}\{1-(1-q_a)^m\} \mid D]. \quad (8)$$

Additional suffixes reduce only that group's residual no-contrast mass. Across eligible groups, the budget problem is

$$\begin{aligned} &\text{maximize } \textstyle\sum_{a\in A_{hom}} U_a(k_a)\\ &\text{subject to } \textstyle\sum_a k_a \;\leq\; B_{Branch},\\ &k_a \;\in\; \{0\} \cup \{4, \ldots, K_a\}. \end{aligned} \quad (9)$$

An opened group receives at least four new suffixes; its factual child costs no new rollout. $K_a$ respects worker capacity and budget. An exact global integer dynamic program selects feasible quotas using the particle estimate of (7). Appendix B.2 derives the event probability and diminishing gains. Actor credit uses realized outcomes after sampling.

## 5. From Physical Comparisons to Actor Updates

### 5.1 Physical segments and terminal backup

Each physical edge is one sampled turn. A child segment E(h,c) starts at split h and ends at the next split or terminal leaf z. Immediate siblings are compared locally (Figure 2). For a terminal leaf, $V_{term}$ equals its verifier reward. For an internal endpoint, it is the mean reward of all descendant leaves L(z):

$$V_{term}(z)=\sum_{\ell\in L(z)} R(\ell)/|L(z)|. \quad (10)$$

This is an observed-tree credit statistic. Endpoint backup weights descendant leaves, while the sibling baseline below weights immediate alternatives equally. Each segment turn receives the endpoint outcome minus its own invalid-action cost $P_i$; the segment score is

$$S(h,c)=(1/|E(h,c)|)\sum_{i\in E(h,c)}[V_{term}(z)-P_i]. \quad (11)$$

For K immediate child segments, trajectory credit is centered across their scores:

$$\mu_h=(1/K)\sum_{c\in C(h)} S(h,c),\quad A_{traj}(h,c)=S(h,c)-\mu_h. \quad (12)$$

The baseline weights immediate children equally, regardless of segment length. Credit stops at the next split; ancestor advantages are not accumulated, and an assigned zero remains zero. Turns without an assignable split use the factual-root fallback in Appendix C.3.

### 5.2 Observation-group step credit

The supporting step term follows GiGPO's observation grouping [4]. Its backup averages immediate outgoing outcomes:

$$V_\gamma(h)=(1/|C(h)|)\sum_{c\in C(h)} T_\gamma(c),\quad T_\gamma(c)=\begin{cases} R(c) & \text{if } c \text{ is terminal} \\ \gamma V_\gamma(c) & \text{otherwise} \end{cases}. \quad (13)$$

We use $\gamma$=0.95 for the step backup in both benchmarks. With $Q_i=T_\gamma(c)-P_i$, exact task/observation groups G(s) center observed turn occurrences, including repeated actions:

$$A_{step,i}=Q_i-(1/|G(s)|)\sum_{u\in G(s)} Q_u,\quad A_i=A_{traj}(h,c)+A_{step,i}. \quad (14)$$

Both raw credit terms have coefficient one. Immediate-child backup and descendant-leaf averaging can differ in unbalanced trees; they define distinct step and segment comparisons.

### 5.3 Complete-turn policy optimization

A response is a complete autoregressive language action. Its score is the sum of its token scores (2). Dividing by its own sampled length introduces the response-dependent factor identified by Dr. GRPO [7]; we retain token sums within each turn and use one active-turn count N for the optimizer minibatch:

$$J_{PG}(\theta)=(1/N)\sum_{i=1}^{N}\sum_{j=1}^{L_i} S_c(\rho_{i,j},A_i),\quad \rho_{i,j}=\frac{\pi_\theta(y_{i,j}|H_i,y_{i,<j})}{\pi_{old}(y_{i,j}|H_i,y_{i,<j})}. \quad (15)$$

$S_c$ is the tokenwise clipped PPO surrogate with a negative-advantage dual cap. At the behavior policy, for a fixed batch and detached advantages, its PG gradient equals $N^{-1}\sum_i A_i \nabla\theta \log \pi_\theta(y_i|H_i)$. This local identity explains the turn-to-token interface; its proof, length-weighting analysis, and limits away from the behavior policy are in Appendix B.3.

PG, reference, and entropy terms share the minibatch's global active-turn count across ranks and partial minibatches; padding is excluded.

The actor adds fixed-reference regularization with a tangent linear tail and a local invalid-syntax cost. Costs enter returns before centering (Appendix D; Appendix B.5).

Unit segment coefficients define the core model. The optional bounded plugin redistributes trajectory coefficient mass across unequal-length sibling segments without changing raw returns, sibling baselines, or step credit (Appendix D). Section 7.2 evaluates this choice.

Algorithm 1 (Appendix D) details a training round. The frozen policy collects Root factual episodes, then Branch suffixes from replayable factual-homogeneous exact task/observation groups under their original deadlines. Completed groups update predictor memory. Immediate physical siblings determine trajectory credit and exact observation groups determine step credit; only turns without assignable splits use the factual-root fallback. The actor sums token surrogates per complete response under the global active-turn denominator. ALFWorld and WebShop share this sequence with benchmark-specific budgets, horizons, and initializations.

## 6. Experimental Design

The actor is Qwen2.5-1.5B-Instruct after environment-specific supervised alignment. Complete ALFWorld evaluation covers 140 seen and 134 unseen tasks, ten stochastic episodes and one greedy episode per task. WebShop covers all 500 test goals in the 1,000-product catalog with six stochastic episodes per goal. The reported ALFWorld and WebShop checkpoints are fixed at updates 175 and 100, respectively, rather than selected as online-monitoring maxima. Each configuration has one training history; task-bootstrap intervals resample evaluation tasks, not training seeds. WebShop online monitoring uses the same goal pool as complete assessment. Appendix A gives decoding and training settings.

## 7. Results

### 7.1 Complete-Task Performance

At the fixed checkpoints, ALFWorld achieves 97.79% seen and 96.64% unseen stochastic success (2,664/2,740 episodes overall).

WebShop achieves 81.07% (2,432/3,000). Table 1 places these complete-task measurements beside published scores, whose initialization and evaluation protocols differ. The two rows report the same allocation method within benchmark-specific training and evaluation configurations.

Table 1. Task-family and overall success (%) for selected RL comparisons. PPO, GRPO, and GiGPO follow [4, Table 1], BiPACE-Q follows [8, Table 4], and HGPO follows [33, Table 1]. Published ALFWorld scores use valid-seen tasks; initializations and evaluation protocols differ. Our ALFWorld rows use ten stochastic episodes per task. The WebShop dagger marks 500 goals, six episodes each, in the 1,000-product catalog.

| Type | Method | ALFWorld | | | | | | | WebShop |
|---|---|---|---|---|---|---|---|---|---|
| | | Pick | Look | Clean | Heat | Cool | Pick2 | All | Succ. |
| RL | PPO [4] | 64.8±3.5 | 40.5±6.9 | 57.1±4.9 | 60.6±6.6 | 46.4±4.0 | 47.4±1.9 | 54.4±3.1 | 51.5±2.9 |
| RL | GRPO [4] | 85.3±1.5 | 53.7±8.0 | 84.5±6.8 | 78.2±7.9 | 59.7±5.0 | 53.5±5.6 | 72.8±3.6 | 56.8±3.8 |
| RL | GiGPO, std [4] | 94.4±5.9 | 67.5±4.6 | 94.8±3.8 | 94.4±7.8 | 79.8±4.7 | 76.4±5.4 | 86.7±1.7 | 65.0±3.2 |
| RL | GiGPO, no std [4] | 96.0±1.4 | 76.5±3.9 | 91.8±5.5 | 91.3±6.3 | 71.7±8.4 | 79.5±7.7 | 86.1±4.7 | 67.4±4.5 |
| RL | HGPO, K=2 [33] | | | | | | | 92.8±1.1 | 71.5±4.0 |
| RL | HGPO, K=4 [33] | | | | | | | 94.85±0.92 | 78.12±2.06 |
| RL | BiPACE-Q [8] | 100.0 | 97.4±3.8 | 100.0 | 100.0 | 96.5±3.6 | 92.0±7.9 | 93.5±1.2 | 71.9±2.1 |
| **RL** | **TSBPO (seen)** | **97.43** | **98.46** | **98.52** | **96.25** | **100.00** | **95.83** | **97.79** | **81.07†** |
| **RL** | **TSBPO (unseen)** | **87.92** | **100.00** | **99.68** | **97.39** | **97.62** | **97.65** | **96.64** | — |

Figures 4(a) and 4(b) show the full ALFWorld and WebShop training histories. Online monitoring and allocation diagnostics describe one history per benchmark; Table 1 reports fixed-checkpoint complete-task evaluations. Both runs use the same fitted allocator with benchmark-specific rollout budgets and horizons.

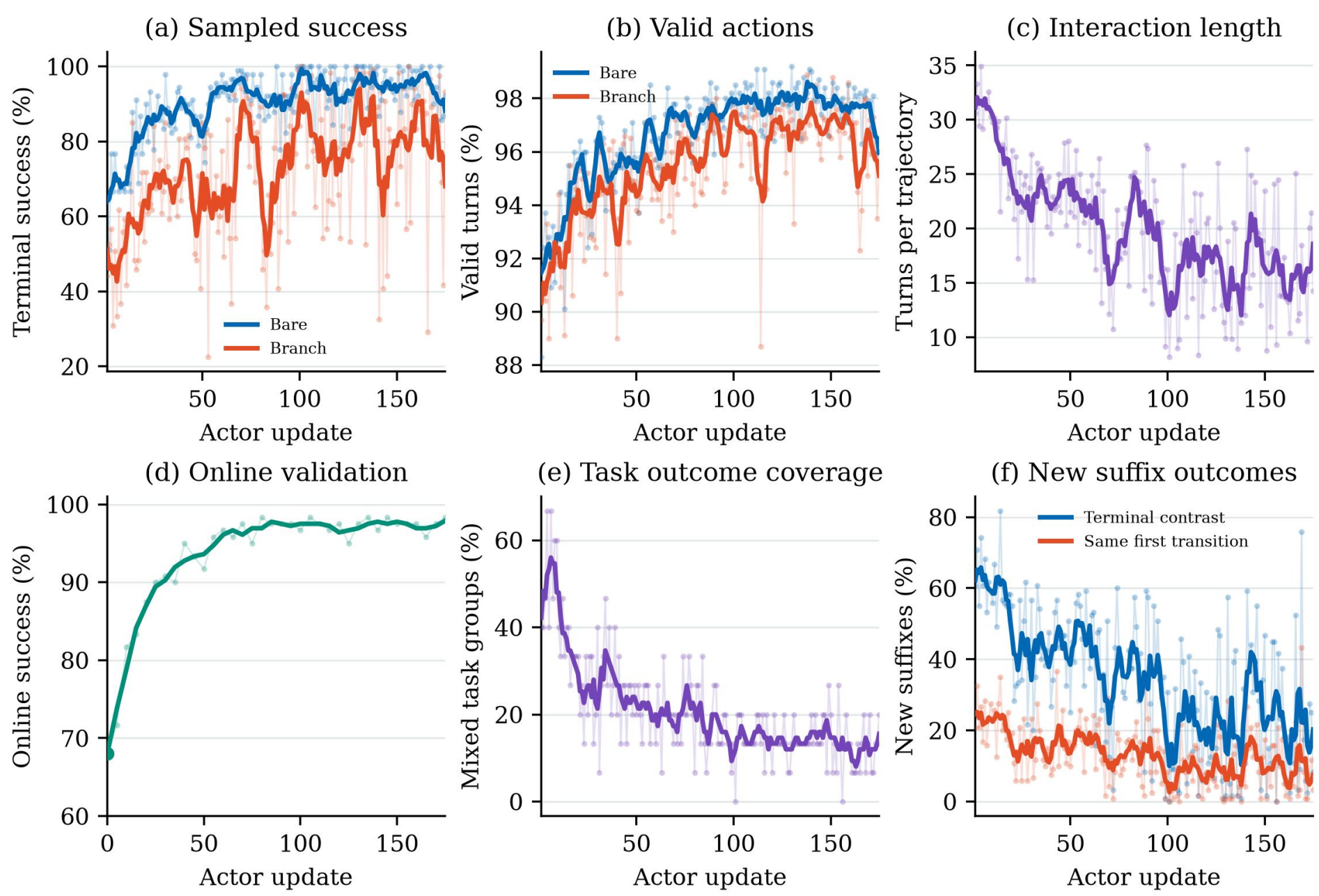

Figure 4(a). Complete ALFWorld training and online-monitoring history through 175 updates. Panels show bare/branch success and action validity, trajectory length, online success, task-group coverage, and new suffix outcomes. Online success samples 120 valid-seen episodes every five updates; complete-task success is reported in Table 1.

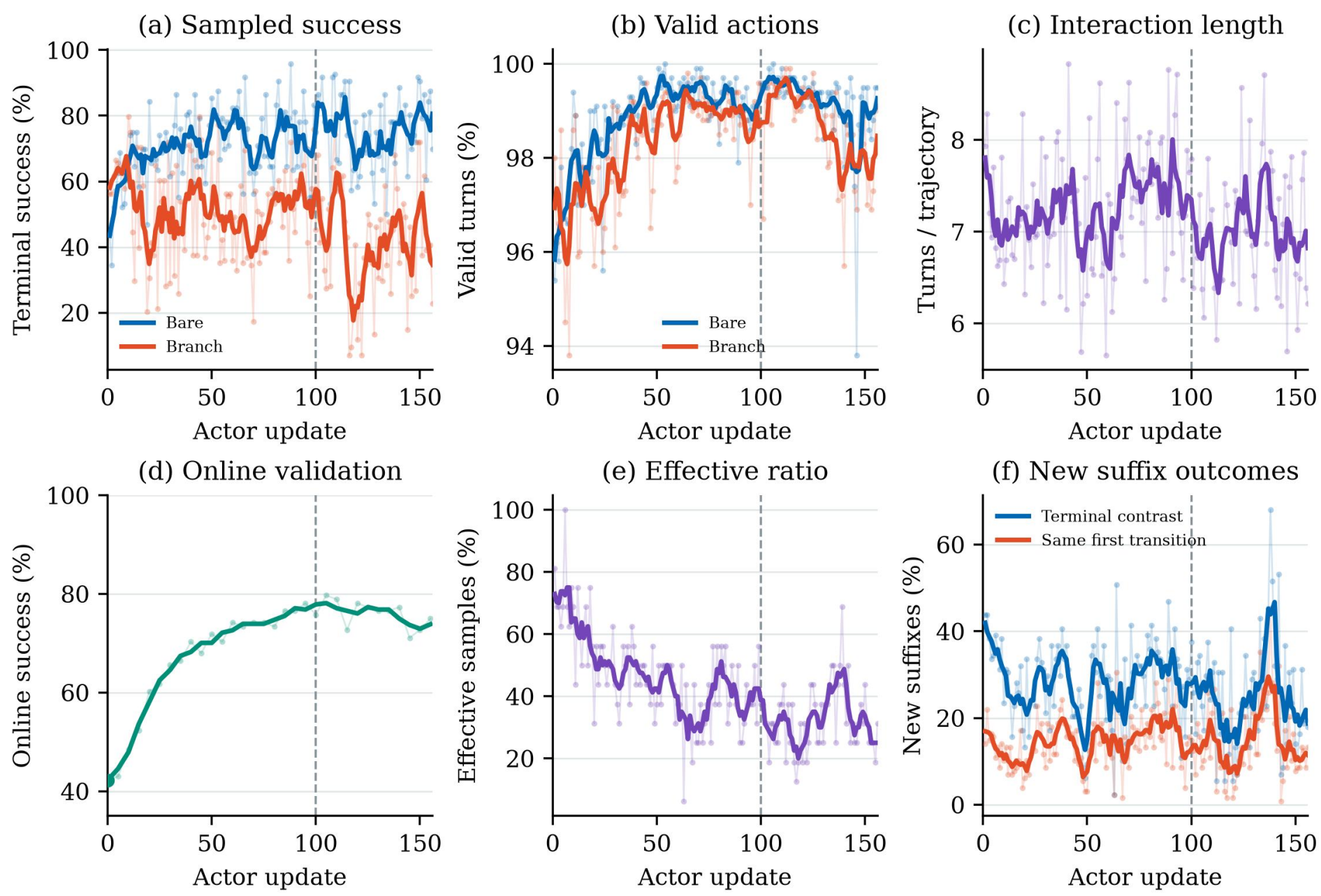

Figure 4(b). Complete WebShop training and online-monitoring history across 156 updates, with the same six panels as Figure 4(a). The dashed line marks the evaluated update-100 checkpoint. Online monitoring samples 128 goals from the pool also used in complete evaluation; updates 101-156 are later training diagnostics.

## 7.2 Actor-update and segment-coefficient comparisons

This ALFWorld comparison changes only policy-gradient aggregation: the default uses loss_agg_mode=seq-mean-token-sum, and the reference uses seq-mean-token-mean. For N active responses of lengths $L_i$, the reference averages the token surrogates within each response before averaging responses:

$$J_{PG}^{norm}(\theta)=N^{-1}\sum_{i=1}^{N}[L_i^{-1}\sum_{j=1}^{L_i}S_c(\rho_{i,j},A_i)].$$

The default removes the response-specific factor $1/L_i$ from the PG term. Appendix B.3 proves that this eliminates the associated length-normalization term under a fixed-context score calculation; it does not prove that adaptive tree advantages are unbiased. Both conditions keep the fixed-reference penalty summed over tokens with the same global active-turn denominator N; entropy regularization remains disabled. Initialization, the Section 4 allocator, rollout budgets, advantages, segment coefficients, optimizer, learning rate, and clipping remain at Table A1 defaults. No learning-rate or regularization compensation accompanies the PG switch. Tables 2-3 summarize the first 100 actor updates (100 records per condition).

Table 2. Interaction-quality means over the first 100 updates of each aggregation condition (100 records). Percentages average training updates; the response-limit monitor is distinct from backend truncation in complete-task evaluation.

| **Diagnostic** | **Aggregation reference** | **TSBPO** |
|---|---|---|
| Valid actions (%) | 92.60 | 96.59 |
| Branch valid actions (%) | 91.60 | 94.59 |
| Valid format (%) | 96.80 | 99.55 |
| Responses at token limit (%) | 2.40 | 0.01 |
| Mean response tokens | 162.18 | 141.80 |

Table 3. Temporal variability and PPO clipping over the same 100 records per condition. SD describes temporal variation rather than variation across independent training seeds.

| **Diagnostic** | **Aggregation reference** | **TSBPO** |
|---|---|---|
| Valid-action SD (pp) | 5.83 | 1.92 |
| Token-limit SD (pp) | 6.03 | 0.03 |
| PPO clipped tokens (%) | 1.81 | 1.31 |

Complete-turn aggregation yields higher observed command validity (96.59% versus 92.60%), fewer responses at the token limit (0.01% versus 2.40%), and shorter mean outputs (141.80 versus 162.18 tokens). With the remaining settings fixed, the aggregation switch changes both the weighting of responses and the magnitude of the PG contribution relative to regularization. These are the measured effects of the specified optimization configuration.

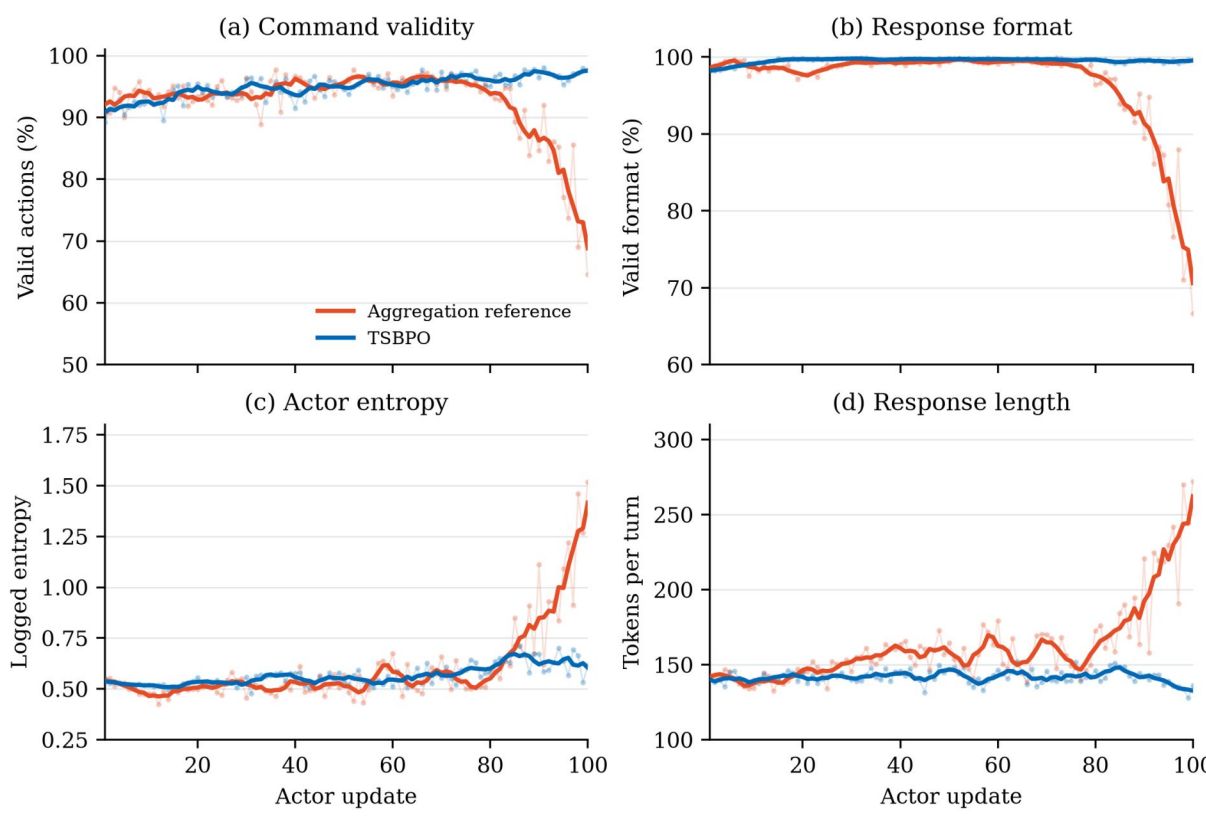

Figure 5. Full ALFWorld PG-aggregation comparison over the first 100 updates. Panels show command validity, response format, actor entropy, and response length. The other actor settings remain fixed; points are measurements and solid lines are centered five-record means.

Relative to Table A1, this ALFWorld comparison changes only the optional segment plugin: unit coefficients versus enabled weighting with cap C=6. Initialization, the Section 4 allocator, rollout budgets, complete-turn aggregation, optimizer, and regularization are identical. Both models receive complete-task evaluation after 175 actor updates. Table 4 reports these outcomes.

Table 4. Complete-task stochastic evaluation of both segment configurations at an equal final training budget. Every model uses the same 140 seen and 134 unseen tasks, ten episodes per task, and the decoding protocol in Appendix A. Overall success weights all 2,740 episodes equally.

| **Configuration** | **Seen success (%)** | **Unseen success (%)** | **Overall success (%)** |
|---|---|---|---|
| Without segment balance | 97.79 | 96.64 | 97.23 |
| With segment balance (C=6) | 96.71 | 94.93 | 95.84 |

Paired measurements on the enabled run's trees show the intended scalar-mass change: the mean worst-sibling relative gap falls from 2.917 before weighting to 1.031 afterward. Bounds are active for 6.47% of child segments. This measures coefficient redistribution, not gradient equality. Appendix Table A5 gives additional bare/branch credit and tree-structure diagnostics.

Unit coefficients achieve higher final stochastic success overall: 97.23% versus 95.84%, a 1.39-point difference with paired task-bootstrap interval −0.26 to 2.96 points. Greedy success also favors unit coefficients, whereas complete-evaluation truncation is lower with the plugin (0.07% versus 0.20%). These single-run comparisons support optional configuration rather than a required performance contribution; their intervals describe evaluation-task variation, not training-seed uncertainty.

### 7.3 WebShop step-advantage ablation

We set the step-advantage coefficient to zero in the actor update, retaining physical-sibling trajectory credit and the Table A2 settings for initialization, allocation, rollout budgets, PG aggregation, segment weighting, optimization, and regularization. The coefficient is one in the full model. Both update-100 checkpoints are evaluated on the same 500 WebShop goals, with six stochastic episodes per goal, temperature 0.4, evaluation seed 0, and a 15-step horizon.

Table 5 reports 81.07% success for TSBPO (2,432/3,000) and 75.93% without step advantage (2,278/3,000). Removing the step term reduces success by 5.13 percentage points; the paired 95% interval for ablation minus full is [−7.80, −2.53] points. Intervals use 20,000 percentile bootstrap resamples of whole goals, preserving all six trials per goal. The full model records more successful trials on 89 goals, the ablation on 49, and both tie on 362.

Table 5. WebShop step-advantage ablation at update 100. Credit magnitudes average the per-update training diagnostics over updates 1–100; success intervals resample evaluation goals.

| **Metric** | **TSBPO** | **w/o step** |
|---|---|---|
| Step coefficient | 1 | 0 |
| Mean absolute step credit | 1.197 | 0.000 |
| Successful episodes | 2,432/3,000 | 2,278/3,000 |
| Success (%)<br>[95% interval] | 81.07<br>[77.97, 84.13] | 75.93<br>[72.33, 79.47] |

Figures 6 and 7 trace the corresponding first 100 actor updates. The disabled step term gives zero local step credit by construction; mixed-outcome rates describe the sampled groups, while online validation tracks the training histories separately from the complete update-100 evaluation in Table 5.

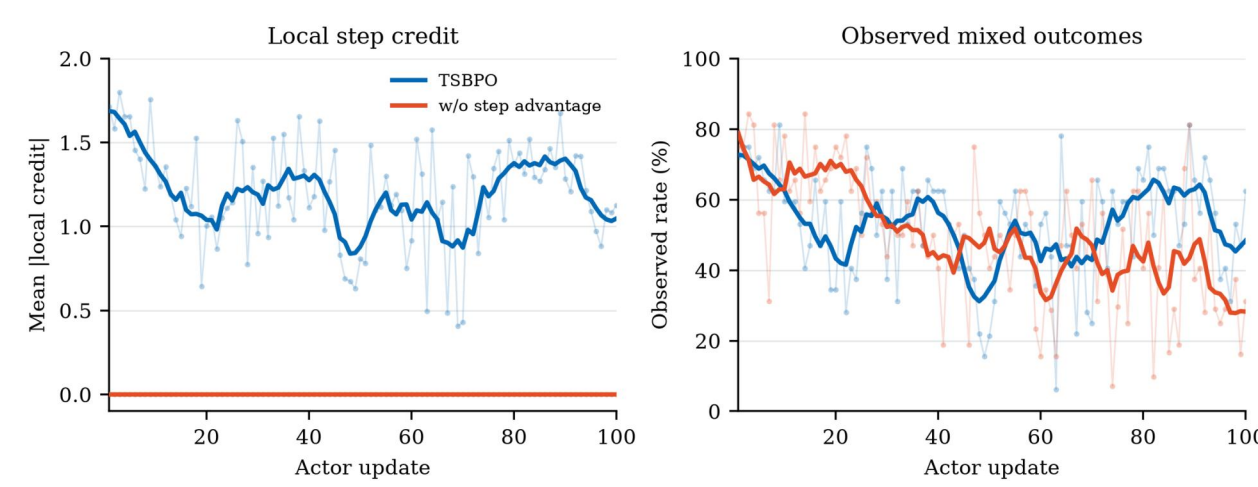

Figure 6. WebShop training diagnostics for TSBPO and the no-step-advantage ablation over updates 1–100. The panels show mean absolute local step credit and observed mixed-outcome rate. Pale points and traces are per-update records; bold curves are centered seven-update means. The outcome rate is a realized sampling diagnostic, not a probability-calibration result.

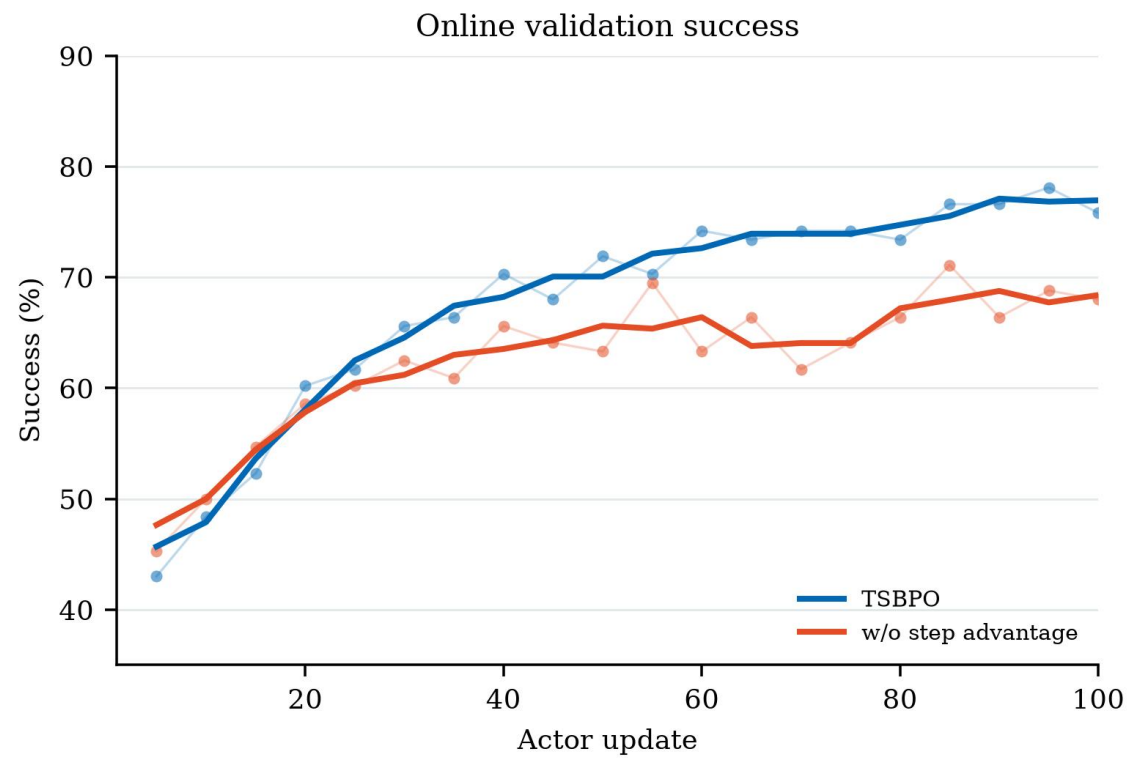

Figure 7. WebShop online-validation success through update 100. Points show recorded validation updates and bold curves are centered three-record means. Each condition has one training run; Table 5 reports the separate complete evaluation of the update-100 checkpoints.

## 8. Discussion

The task evaluations establish performance of the two stated training configurations; suffix outcomes show the alternatives they acquired. The allocator chooses samples before credit is computed, while observed sibling and observation-group outcomes determine the subsequent advantages. Online prediction reliability is a separate property from preferentially sampling high-scored groups, and the WebShop diagnostics show measurable overprediction. The WebShop step-term comparison favors the full update at the evaluated checkpoint.

## 9. Limitations

Each configuration has one training run. Task-bootstrap intervals do not measure training-seed variation. WebShop monitoring and complete evaluation share the same goal pool. Adaptive tree means are observed-tree credit proxies, without a general unbiasedness, monotonic-improvement, or convergence result. The two benchmarks and one actor scale do not establish broader generality.

## 10. Conclusion

TSBPO connects outcome-contrast sampling, physical-sibling and observation-group credit, and complete-turn updates for interactive language agents. The evaluated configurations achieve 97.79% seen and 96.64% unseen ALFWorld success and 81.07% WebShop success; WebShop's likelihood-fitted predictions also show online overprediction.

AI assistance disclosure. A large language model assisted with drafting and editing this manuscript; the authors are responsible for the final text, analyses, and references.

# Appendix A. Models, Configuration, and Evaluation Details

## Model, Training, and Evaluation Configuration

For ALFWorld and WebShop, Qwen2.5-1.5B experiments are run on 6×RTX 4090 GPUs. Tables A1 and A2 give the configurations of the evaluated ALFWorld and WebShop models. Both start from environment-specific SFT of Qwen2.5-1.5B-Instruct [15] and use that initialization as the fixed RL reference. Both evaluated models use the likelihood-fitted Root and Branch path with bounded semantic evidence, original remaining deadlines, and local-state-contrast Branch utility (Section 4).

Table A1. Default ALFWorld configuration for all experiments unless a comparison explicitly specifies a change. Root direct evidence decays by 0.95 per round; Branch-state evidence decays by 0.90.

| Parameter | Setting |
|---|---|
| Actor / fixed reference | Environment-specific Qwen2.5-1.5B SFT |
| Allocator | Likelihood-fitted Root/Branch; local-state contrast |
| Selected tasks / candidate pool | 15 / 120 |
| Root / additional suffix budgets | 90 / 120 |
| Minimum roots / new suffixes per opening | 2 / 4 |
| Rollout temperature / top_p / top_k | 1.0 / 1.0 / −1 |
| PG aggregation | loss_agg_mode=seq-mean-token-sum |
| Actor learning rate | $2\times10^{-6}$ |
| PPO minibatch / per-rank microbatch | 240 / 4 |
| Reference / entropy coefficients | 0.05 / 0 |
| Root direct evidence decay | 0.95 |
| Branch-state evidence decay | 0.90 |
| Episode action limit / history turns | 50 / 3 |
| Prompt / generated-token limits | 2,048 / 512 |
| Optional segment plugin | Disabled (unit coefficients) |
| Nominal root group / prediction group | 6 / 6 |
| Task-specific root quota cap | None (full root budget) |
| Branch worker / per-anchor capacity | 64 / 64 new suffixes |
| Optimizer / PPO epochs per rollout batch | AdamW / 1 |
| Environment / generation / data-sampler seeds | 0 / 0 / 1 |
| Allocation / actor-row permutation seeds | 17 / 17 + 1,000,003 × update |

The complete ALFWorld assessment starts from every seen and unseen validation task. Each task has ten stochastic episodes at temperature 0.4, top_p=1, top_k=−1, and repetition penalty one, plus one greedy episode. Episodes have at most 50 environment steps, 512 generated tokens per turn, three turns of history, and a 2,048-token prompt limit. Stochastic success is the fraction of successful episodes; the task-bootstrap interval resamples whole tasks with their ten outcomes. Action validity requires format compliance and command admissibility, and backend truncation follows the generation termination record.

ALFWorld online monitoring runs 120 stochastic episodes sampled from the valid-seen split every five actor updates; the denominator is episodes, and task identities need not be distinct. Decoding uses temperature 0.4, top_p=1, top_k=−1, and repetition penalty one, with the same 50-action limit, three history turns, 2,048-token prompt limit, and 512-token response limit as complete-task assessment. The approximately 68% initial anchor uses this smaller monitoring sample, whereas the approximately 66% independent SFT result covers all seen tasks with repeated episodes.

Table A2. Evaluated WebShop likelihood-fitted configuration through update 100.

| Parameter | Setting |
|---|---|
| Actor / fixed reference | Environment-specific Qwen2.5-1.5B SFT |
| Allocator | Likelihood-fitted Root/Branch; local-state contrast |
| Selected tasks / candidate pool | 16 / 128 |
| Root / additional suffix budgets | 96 / 128 |
| Minimum roots / new suffixes per opening | 2 / 4 |
| Rollout temperature / top_p / top_k | 1.0 / 1.0 / −1 |
| PG aggregation | loss_agg_mode=seq-mean-token-sum |
| Actor learning rate | $2\times10^{-6}$ |
| PPO minibatch / ppo_max_token_len_per_gpu | 240 / 5120 |
| Reference / entropy coefficients | 0.05 / 0 |
| Root direct evidence decay | 0.95 |
| Branch-state evidence decay | 0.90 |
| Episode action limit / history turns | 15 / 3 |
| Prompt / generated-token limits | 4,608 / 512 |
| Optional segment plugin | Enabled, cap C=6 |
| Nominal root group / prediction group | 6 / 6 |
| Task-specific root quota cap | None (full root budget) |
| Branch worker / per-anchor capacity | 64 / 64 new suffixes |
| Optimizer / PPO epochs per rollout batch | AdamW / 1 |
| Environment / generation / data-sampler seeds | 0 / 0 / 1 |
| Allocation / actor-row permutation seeds | 17 / 17 + 1,000,003 × update |

Complete WebShop assessment uses the 1,000-product catalog and all 500 synthetic test goals, indexed 0-499 under goal-order seed zero. Each receives six stochastic episodes at temperature 0.4, top_p=1, top_k=-1, with at most 15 actions, three history turns, 4,608 prompt tokens, and 512 generated tokens per turn. The environment's won flag determines success. Task-bootstrap intervals use 10,000 whole-task resamples, preserving six attempts per goal. Online monitoring samples 128 distinct goals from the same 500-goal test pool every five updates, with one stochastic episode per goal and the same decoding and interaction limits; training samples goals outside that pool. These online scores track training but do not rank saved checkpoints for the reported complete assessment. The assessment evaluates the update-100 model on the same goal pool with additional episodes, rather than on a disjoint task set.

Table A3. Additional complete WebShop evaluation diagnostics for the update-100 model. The 20,104 evaluated turns include invalid outputs but exclude padding; turn rates use this denominator. Horizon termination counts episodes.

| Diagnostic | Count or value | Denominator / unit |
|---|---|---|
| Evaluated interaction turns | 20104 | Turns |
| Format-valid outputs | 99.99% | Evaluated turns |
| Executable actions | 99.88% | Evaluated turns |
| Mean generated response | 113.19 | Tokens / turn |
| Unsuccessful episodes | 568 | 3000 episodes |
| Failed episodes at horizon | 256 | 3000 episodes |
| All episodes at horizon | 260 | 3000 episodes |
| Truncated outputs | 3 | Turns |
| Mean successful / failed length | 5.84 / 10.41 | Actions / episode |

External protocols. HGPO’s K=2 and K=4 settings denote historical context depth, not rollout count; both start from Qwen2.5-1.5B-Instruct and allow 30 WebShop actions [33, Sections 4.2 and C.3]. BiPACE reports maximum aggregate validation success over training; its ALFWorld family entries can select different checkpoints [8, Table 4 and Appendix N]. Our rows evaluate one model per benchmark at the checkpoint specified in Section 7.1. These protocols determine the contextual role of the published scores in Table 1.

## SFT Cold-Start Data Collection and Configuration

Before online RL, successful interactions generated by the GPT-5.6-sol teacher align the actor with the environment’s observation, reasoning, and action syntax. Both benchmarks use the same collection procedure: the teacher interacts with training environments, successful complete trajectories are retained, and each retained turn becomes a user–assistant pair in ShareGPT format. The user message contains the environment prompt; the assistant message contains the generated response, with a <think>...</think> block followed by an <action>...</action> block. Supervised training applies causal next-token likelihood to assistant tokens. The resulting environment-specific model initializes both the RL actor and its fixed reference.

Table A4 gives the supervised cold-start settings. For both benchmarks, the student starts directly from Qwen2.5-1.5B-Instruct and undergoes two epochs of full-parameter SFT on the corresponding successful-trajectory dataset. Both use the Qwen chat template, DeepSpeed ZeRO-2, BF16, FlashAttention 2, gradient checkpointing, and cosine learning-rate decay. The five-percent held-out split comes from each training dataset and monitors supervised loss; it is separate from environment validation tasks.

Figures A1 and A2 give the history-conditioned turn prompts. Braced fields are filled at runtime; colors distinguish semantic slots and response tags.

**Prompt Template for ALFWorld**

You are an expert agent operating in the ALFRED Embodied Environment. Your task is to: {task_description} Prior to this step, you have already taken {step_count} step(s). Below are the most recent {history_length} observations and the corresponding actions you took: {action_history} You are now at step {current_step} and your current observation is: {current_observation} Your admissible actions of the current situation are: [{admissible_actions}].

Now it's your turn to take an action. You should first reason step-by-step about the current situation. This reasoning process MUST be enclosed within <think> </think> tags. Once you've finished your reasoning, you should choose an admissible action for current step and present it within <action> </action> tags.

Figure A1. ALFWorld turn prompt. Braced fields denote runtime inputs; the response uses the displayed reasoning/action tags.

**Prompt Template for WebShop**

You are an expert autonomous agent operating in the WebShop e-commerce environment. Your task is to: {task_description}. Prior to this step, you have already taken {step_count} step(s). Below are the most recent {history_length} observations and the corresponding actions you took: {action_history} You are now at step {current_step} and your current observation is: {current_observation}. Your admissible actions of the current situation are: [ {available_actions} ].

Now it's your turn to take one action for the current step. You should first reason step-by-step about the current situation, then think carefully which admissible action best advances the shopping goal. This reasoning process MUST be enclosed within <think> </think> tags. Once you've finished your reasoning, you should choose an admissible action for current step and present it within <action> </action> tags.

Figure A2. WebShop turn prompt. Braced fields denote runtime inputs; the response uses the displayed reasoning/action tags.

The WebShop collection pipeline uses the 1,000-product training catalog, excluding sessions 0–499 and their associated validation queries. Queries with insufficient successful demonstrations receive priority. Up to eight environment sessions and sixteen teacher requests run concurrently. The prompt includes the goal, current page, admissible commands, and at most five recent observation–action pairs. Teacher requests allow 2,048 completion tokens and 512 tokens for reflection, use high reasoning effort, and leave temperature and top_p at the service defaults. The collector checks the two tagged blocks and command admissibility, allowing two repairs before execution.

Table A4. Supervised cold-start settings. Both students start directly from Qwen2.5-1.5B-Instruct and train for two epochs; demonstrations are collected with GPT-5.6-sol.

| Setting | ALFWorld | WebShop |
|---|---|---|
| Example format | User–assistant pairs | User–assistant pairs |
| Training examples | 3,040 | 4,605 |
| Prompt cutoff (tokens) | 2,048 | 5,120 |
| Training batch per device | 4 | 2 |
| Gradient accumulation | 2 | 4 |
| Learning rate | $5\times10^{-6}$ | $5\times10^{-6}$ |
| Training epochs | 2 | 2 |
| Warmup ratio | 0.10 | 0.05 |
| Supervised validation split | 0.05 | 0.05 |
| Evaluation batch per device | 4 | 2 |
| Evaluation interval (updates) | 200 | 200 |
| Logging interval (updates) | 10 | 10 |

Each shopping attempt allows fifteen actions. A failed attempt can supply private reflection to a subsequent attempt of the same query, with at most three attempts. Only trajectories whose terminal environment flag indicates success enter the exported dataset. Their original user prompts and final valid assistant responses are saved as one pair per turn; private reflection and repair messages are excluded. The WebShop supervised dataset contains 4,605 user–assistant pairs from successful training trajectories.

The environment-specific BF16 supervised models initialize both the RL actor and fixed reference. Historical SFT assessments retain an earlier assessment protocol: two episodes on each ALFWorld seen task with Q8_0 inference (185/280 successes), and 512 WebShop episodes (266 successes). It supplies initialization context; differences from the RL rows are not reported as matched SFT-to-RL gains.

## Tree Structure and Credit-Composition Diagnostics

Table A5 reports tree-structure and sampling diagnostics used in Section 7.2. For a sibling group with child masses $m_c$, its relative gap is $(\max_c m_c - \min_c m_c)/\mathrm{mean}_c\, m_c$; each training update reports the maximum across groups. Coefficient quantiles are computed over active turn rows, whereas cap frequency counts child segments. Absolute combined advantage averages the magnitude of the composed row advantage separately for bare and branch rows. A sign conflict means the product of the trajectory and state-step components is negative. These are distinct measures of tree weighting and credit composition, rather than independent-trajectory counts or direct gradient-variance estimates.

Table A5. Tree coefficient masses and credit composition. ALFWorld columns average 175 updates per run; the evaluated WebShop column averages updates 1-100 of the likelihood-fitted run. Each diagnostic contributes one record per update. Before/after values within an enabled run use the same tree batch; cross-run differences also include policy and sampled-tree variation. WebShop provides weighting context rather than an on/off control.

| Diagnostic | ALFWorld plugin off | ALFWorld plugin on | WebShop plugin on |
|---|---|---|---|
| Worst sibling mass gap before weighting | 2.851 | 2.917 | 2.575 |
| Worst sibling mass gap after weighting | 2.851 | 1.031 | 0.360 |
| Child segments at coefficient bounds (%) | 0.00 | 6.47 | 1.61 |
| Median active-row coefficient | 1.000 | 0.878 | 0.935 |
| 95th-percentile active-row coefficient | 1.000 | 1.864 | 1.589 |
| Branch fraction of active rows (%) | 64.15 | 65.61 | 55.99 |
| Absolute combined advantage, bare | 1.935 | 2.126 | 2.488 |
| Absolute combined advantage, branch | 4.290 | 4.127 | 3.415 |
| Trajectory/state sign conflict, bare (%) | 18.12 | 14.68 | 9.59 |
| Trajectory/state sign conflict, branch (%) | 13.87 | 11.92 | 7.77 |

## Task-Group Sampling-Ratio Definition

Panel (e) of Figure 4(a) and Figure 4(b) reports the task-group effective sampling ratio using the update-level definition below.

$$e_r = |T_r|^{-1} \sum_{t \in T_r} 1[0 < \sum_{b \in B_{t,r}} Y_b < |B_{t,r}|].$$

Here $T_r$ is the set of sampled task identities, $B_{t,r}$ contains distinct completed terminal trajectories after branching, and $Y_\beta$ is binary success; duplicate turn rows do not create new trajectories. An empty task set has ratio zero. Suffix-contrast rates divide the number of qualifying new suffixes by max(1, number of completed new suffixes); selected-group conversion rates use the same max(1, denominator) convention for eligible groups with completed evidence. Thus a zero-denominator update contributes zero to the unweighted update mean. Candidate and selected probability means summarize representatives before branching.

# Appendix B. Conditional Predictor and Optimizer Derivations

## B.1 Predictive mixed-group probability

For a fixed success probability p and $n \geq 1$ conditionally independent outcomes from one task, the probability of observing both labels is $1-p^n-(1-p)^n$. Within a fitted mixture component $p \sim$ Beta(A,B), with A,B>0, the beta integral gives

$$E[p^n] = B(A+n,B)/B(A,B),$$
$$E[(1-p)^n] = B(A,B+n)/B(A,B). \quad \text{(B.1)}$$

Subtracting the two all-equal events gives the component utility u(n;A,B). Averaging u over the task-specific weights $w_i(\theta)$ yields the Root mixture utility (5). Because these weights do not depend on the proposed n, its one-rollout gain is

$$U(n+1)-U(n) = E[p^n(1-p)+(1-p)^n p] \geq 0. \quad \text{(B.2)}$$

Both terms in the integrand decrease as n increases, so each component and their weighted mixture have diminishing gains after the first rollout. The zero-to-one gain is zero; Root allocation therefore starts each selected task with two factual rollouts. This is a conditional predictive calculation. Age-weighted evidence and policy drift do not establish that the resulting mixture is a calibrated probability of future success.

## B.2 State-group contrast and its marginal gains

Condition on evidence D and the modeled opposite-outcome probability $q_a$ for a factual-homogeneous state group a. If new suffix labels are conditionally independent given $q_a$, the chance that all $k_a$ new suffixes retain the factual label is $(1-q_a)^{k_a}$. Subtracting this chance from one and integrating over the predictive law proves (7). Linearity of expectation makes the sum in (9) the expected number of newly mixed factual-homogeneous groups, even when the predictive particles are correlated across groups. Subtracting the local no-contrast probabilities before and after m additional suffixes proves (8).

For the same state group, let $l \geq k$ be two suffix quotas. Since $0 \leq 1-q_a \leq 1$, the residual no-contrast factor at l is no larger than at k in every particle. Thus the same added block has

$$0 \leq \Delta_{a,m}(l) \leq \Delta_{a,m}(k). \quad \text{(B.3)}$$

For fixed $k_a$, the gain per added suffix is nonincreasing in the block size. These statements hold for a fixed predictive law supported on [0,1] and for its finite particle average. They do not prove that the particle law is calibrated to future actor outcomes. The minimum opening quota rules out an unrestricted greedy claim, but the integer dynamic program exactly maximizes the finite-particle objective over the implemented feasible quotas.

The counted event is a new terminal-label contrast at a factual-homogeneous state group. One task can contribute several eligible groups, including when its factual trajectories are already mixed across other observations. A newly mixed observation group need not create a mixed group at every physical sibling split, and local costs can create score differences even when labels agree. Acquisition therefore directs sampling; it does not guarantee a nonzero actor gradient.

## B.3 Complete-turn aggregation and length normalization

For the local score identity in Section 5.3, condition on a fixed batch with contexts, generated tokens, N>0 active turns, and detached advantages. Sampled tokens have positive behavior probability and differentiable policy probabilities near $\theta_{old}$. For ε>0, each ratio equals one strictly within its clipping interval. For negative A and dual cap c>1, A>cA, so the cap is locally inactive. Zero advantages contribute zero. At the behavior policy,

$$\nabla_\theta S_c(\rho_{i,j},A_i)=A_i\nabla_\theta \log\pi_\theta(y_{i,j}|H_i,y_{i,<j}). \quad (B.4)$$

Summing over tokens and differentiating the autoregressive factorization (2) yields

$$\nabla_\theta J_{PG}(\theta_{old})=(1/N)\sum_i A_i\nabla_\theta \log\pi_\theta(y_i|H_i)|_{\theta=\theta_{old}}. \quad (B.5)$$

Equation (B.5) is an exact fixed-batch gradient identity; it does not assert that the sampled advantages equal unbiased policy advantages. EOS is included only when generated, and padding contributes nothing. Deterministic detached row weights enter the sum in the same way. The denominator is the sampled active-turn count, not an inverse episode-inclusion probability; episodes with more retained turns contribute more terms before optional reweighting.

For the response-length result, fix a context H and suppose the response law has differentiable probabilities with common support and positive finite lengths L(y). Let R(y) have no direct dependence on θ and let b=b(H) be a finite baseline independent of the sampled response. Interchanging differentiation and expectation, the score identity gives E[(R-b)∇θ log πθ(y|H)]=∇θ E[R]. Dividing each response score by its sampled length instead gives E[((R-b)/L(y))∇θ log πθ(y|H)]=∇θ E[R/L(y)]-b∇θ E[1/L(y)]. In particular,

$$E[(b/L(y))\nabla_\theta \log\pi_\theta(y|H)]=b\nabla_\theta E[1/L(y)]. \quad (B.6)$$

The undivided baseline term vanishes because E[∇θ log πθ(y|H)]=∇θ1=0; the length-divided term in (B.6) need not vanish. For a concrete example, suppose the only responses have lengths one and two, the second has probability q(θ), R(y)=0 for both, and b>0. The expected undivided, baseline-subtracted PG score is zero, while the length-divided, baseline-subtracted score is (b/2)∇θ q(θ), nonzero wherever q varies. Thus removing 1/L(y) exactly removes this normalization-induced mechanism, as in Dr. GRPO [7]. It does not establish unbiasedness for TSBPO's empirical sibling baselines or outcome-dependent tree selection.

Away from $\theta_{old}$, the complete-response likelihood ratio is $\prod_j \rho_{i,j}$; a sum of tokenwise clipped terms generally differs from a surrogate based on that product. The local score identity applies to the PG component before reference regularization, not to exact sequence importance sampling across multiple PPO updates.

### B.4 Bounded scalar-mass projection

Consider K positive segment masses $m_c$, total $T=\sum_c m_c>0$, mean $\bar{m}=T/K$, and C≥1. Define $a_c=\bar{m}/m_c$ and $F(\lambda)=\sum_c m_c \mathrm{clip}(\lambda a_c,1/C,C)$. For C>1, F is continuous and nondecreasing, $F(0)=T/C\le T$, and at $\lambda_{hi}=C/\min_c a_c$ all factors reach C, giving $F(\lambda_{hi})=CT\ge T$. Here every $a_c$ is positive, so the denominator is positive. The intermediate value theorem establishes a solution to

$$F(\lambda)=T,\lambda\in[0,C/\min_c a_c]. \quad (B.7)$$

Bisection moves its lower bound upward when F<T and its upper bound downward otherwise. Sixty iterations give the projection used by the actor, up to numerical precision. For C=1 every coefficient is one. If no bound is active, $F(\lambda)=\lambda K\bar{m}=\lambda T$, hence λ=1 and every effective child mass equals $\bar{m}$. Active bounds preserve total mass while permitting unequal child masses.

Zero-weight rows and empty responses do not enter active policy groups, so their segments have no positive policy mass to balance. For a formal zero-mass segment its coefficient may be set to one and its contribution is zero; an all-zero group can be omitted. Equalization concerns positive masses only. Neither mass conservation nor equality determines parameter-gradient norms, which also depend on token score vectors and advantages.

### B.5 Reference penalty

For δ≤δ*=log(1+M), the derivative of (D.2) is exp(δ)−1; above the join it is M. Both the value M−δ* and derivative M agree at the join. The derivative is nondecreasing, and δ=0 is a global minimum with value zero. Thus $k_M$ is nonnegative, convex, and continuously differentiable. Its tangent continuation retains a correcting derivative without exponential growth. The modified tail defines a regularizer rather than an exact KL identity.

## Appendix C. Predictive and Fallback Definitions

### C.1 Likelihood fitting from complete terminal counts

Root and Branch fit separate finite mixtures over θ=(a,b,η,ρ). The base success probability has Beta(a,b) shape; η and ρ scale historical direct and capped semantic evidence, respectively. For a query with historical counts ($h_s$, $h_f$), current bare counts ($c_s$, $c_f$), and semantic masses ($m_s$, $m_f$), the component shapes are $A_\theta=a+\eta h_s+c_s+\rho m_s$ and $B_\theta=b+\eta h_f+c_f+\rho m_f$. Root has $c_s=c_f=0$; Branch conditions once on completed current bare outcomes.

The grid combines 25 base means from 0.025 to 0.975, 31 logarithmically spaced base concentrations from 0.001 to 100, and η,ρ∈{0,0.03125,0.125,0.5,1}, for 19,375 equally weighted initial components. An audited physical group g with $s_g$ successes and $f_g$ failures contributes the beta-binomial count likelihood

$$L_g(\theta)=\binom{s_g+f_g}{s_g}\frac{B(A_g(\theta)+s_g,B_g(\theta)+f_g)}{B(A_g(\theta),B_g(\theta))}. \quad (C.1)$$

For query task t, population fitting excludes that task and uses complete groups available before allocation. A group collected at round $r_g$ receives fitting weight $c_g=\lambda^{r-r_g}/M_{t_g,r_g}$, where $M_{t_g,r_g}$ counts groups from its task in that round; λ is 0.95 for Root and 0.90 for Branch. Its preallocation evidence snapshot determines $A_g$ and $B_g$ in (C.1). Semantic masses are covariates, not additional terminal labels.

With a uniform prior over the finite grid, the normalized, task-excluded component weights satisfy

$$w_t(\theta)\propto\exp[\sum_{g:t_g\neq t}c_g\log L_g(\theta)]. \quad (C.2)$$

At least two other tasks are required to fit population weights; otherwise the prediction uses Beta(1/2,1/2). Without identifiable historical evidence η remains one, and without identifiable semantic evidence ρ is zero. When Branch groups are sparse, complete factual Root groups from other tasks can initialize its base distribution, but Root reliability coefficients are not transferred. Only a group whose complete physical outcome count equals its assigned quota enters likelihood fitting. The age-weighted conditional likelihood is a drift-aware predictive approximation, not an exact stationary hierarchical posterior.

## C.2 Shared Branch prediction particles

For each task t with eligible state groups and particle index ℓ, sample one $\theta_t^{\ell}$ from its task-excluded Branch mixture and share that component across all candidate anchors of t. Draw task-shared base masses $G_{t,+}^{\ell}\sim$ Gamma($a_\theta$,1) and $G_{t,-}^{\ell}\sim$ Gamma($b_\theta$,1). A physical sampling cluster b contains $n_b$ paths; in state/depth pool a, its compressed counts $s_{ab}$ and $f_{ab}$ record the successful and failed paths visiting a once, while $u_{ab}=n_b-s_{ab}-f_{ab}$ counts the other paths. Equal visiting-subset fingerprints reuse the same within-cluster draw.

For particle ℓ, the pool/cluster scale is $z_{ab}^{\ell}=0.90^{age}\,\eta_{\theta_t^{\ell}}$ for historical direct evidence, one for current bare evidence, and the bounded semantic donor factor times $\rho_{\theta_t^{\ell}}$ for borrowed evidence. Draw a cluster total with Gamma($n_b\, z_{ab}^{\ell}$,1) shape and visiting-category gamma masses with shapes ( $s_{ab} z_{ab}^{\ell}$,$f_{ab} z_{ab}^{\ell}$,$u_{ab} z_{ab}^{\ell}$); zero-count categories have zero mass. Normalize category masses to the shared total. The total is reused for the same cluster and scale vector, preserving dependence between matching compressed subsets.

$$X_{a,\pm}^{\ell}=G_{t,\pm}^{\ell}+\sum_b W_{ab,\pm}^{\ell}, p_a^{\ell}=X_{a,+}^{\ell}/(X_{a,+}^{\ell}+X_{a,-}^{\ell}). \quad (C.3)$$

Here $W_{ab,\pm}^{\ell}$ is the scaled successful or failed category mass from cluster b. Historical scale includes original collection age and the particle's η; current bare uses unit scale, while semantic donor scale includes retrieval similarity, age, caps, and the particle's ρ. An exact-query cluster cannot also be a semantic donor, and each donor cluster supplies at most one visiting subset. Shared fingerprints capture equal subsets; partial overlaps between distinct compressed subsets remain an approximation. Equation (C.3) gives one success-probability draw for the original deadline H=L−d of pool a.

The 512 aligned indices form a joint vector across anchors of a task. For factual-homogeneous state groups, each $q_a^{\ell}$ is $p_a^{\ell}$ after factual failure or $1-p_a^{\ell}$ after factual success, and the particle average evaluates (7). A task already mixed across its factual trajectories can still have eligible homogeneous state groups. Moment-matched scalar Beta parameters are diagnostics rather than independent laws used by the budget solver. Only complete audited suffix groups fit the Branch prediction model; valid completed physical suffixes update state memory without crediting replayed ancestors. The particle law approximates shared task and trajectory-cluster dependence and does not establish calibrated future success probabilities by itself.

## C.3 Factual-root compatibility credit

Let $B_x$ be the distinct factual root trajectories for task x and $F_b$ the active turns belonging to factual root b. Conditional branch suffixes are not additional members of $B_x$. With terminal return $R_b$ and local cost $P_i$, define

$$S_b^{root}=(1/|F_b|)\sum_{i\in F_b}(R_b-P_i), \quad A_b^{root}=S_b^{root}-(1/|B_x|)\sum_{b'\in B_x}S_{b'}^{root}. \quad (C.4)$$

An active turn without an assignable physical sibling segment inherits $A_b^{root}$ from its factual root. Assigned local credit takes precedence, including when zero; the fallback is not added. The observation-group step term is composed as in (14). With the optional plugin, fallback turns of the same root form a child segment of a task-root group and use the same scalar-mass rule.

## C.4 Semantic evidence and allocation settings

A frozen Qwen3-Embedding-0.6B encoder maps each task instruction to an L2-normalized vector $e_i$; observations, prompt templates, and interaction histories do not enter this Root feature. Among other tasks with direct outcomes, at most eight neighbors with cosine similarity at least 0.95 receive kernel weight $[\max(0,(\cos(e_i,e_j)-0.95)/0.05)]^2$. Each source success/failure pair is first capped at total mass six; the weighted donor sum is capped at total mass four. The resulting semantic masses ( $m_i^{s}$, $m_i^{f}$) enter each Root component with its learned $\rho_\theta$ as in Section 4.2. They remain query-only covariates, never physical terminal observations or recursively borrowed evidence.

Branch state pools query bounded semantic counts from similar historical states. The same frozen encoder produces normalized features for the task goal and for the goal together with the current observation. Encoded text preserves numbers, negations, prices, and object attributes. A donor must have the same depth and episode horizon, state cosine similarity above 0.95, and goal cosine similarity above 0.90. At most eight eligible historical pools are inspected in descending state similarity. For state and goal similarities $c_s$ and $c_g$, the donor kernel is $\kappa=[(c_s-0.95)/0.05]^2[(c_g-0.90)/0.10]^2$.

Only physical terminal counts from earlier rounds can be borrowed. A sampling cluster already present in the exact query pool is excluded; each remaining cluster contributes at most one visiting subset from its highest-ranked eligible donor. That subset is rescaled to mass at most one, weighted by κ and its original 0.90 age decay, and included in a globally capped semantic mass. The configured raw borrowed-mass cap is four for ALFWorld. The particle-specific $\rho_\theta$ scales this mass during prediction. Borrowed masses are not committed as direct outcomes, and physical state groups remain exact task/observation groups.

The Root curriculum scores the integrated mixture event (5) at $n_{ref}=\max(1, n_{rollout})$, with $n_{rollout}=6$ in both benchmarks; a selected task receives at least two factual rollouts. The remaining Root budget follows the same curve's marginal gains, and no task-specific upper cap is supplied in the default configurations. Branch opening requires four new suffixes per anchor, capped by the available budget and worker capacity; its factual child is retained separately. Historical state memory is bounded at 100,000 entries. Branch uses 512 aligned particles and an exact global budget dynamic program. Tie breaking derives from seed 17, the round, and stable pool identity; 17 is not the global training seed.

# Appendix D. Actor Configuration and Training Procedure

**Algorithm 1 Dynamic curriculum learning and TSBPO**

**Require: actor $\pi_\theta$, fixed reference $\pi_{ref}$, memory $M$, budgets $B_{root}$,$B_{branch}$, optional balance flag and cap $C$.**
**for each training iteration do**
  Set $\pi_{old} \leftarrow \pi_\theta$; advance the round and decay historical evidence in M.
  *// Dynamic curriculum and complete multi-turn rollout*
  Assemble fresh and revisited task candidates; score each by integrated Root mixed-outcome probability and sample with exploration (4)–(5).
  Reserve two factual rollouts per selected task; allocate the remaining Root budget by predicted marginal gains (5).
  **for each selected task $x$ and its root quota $n_x$ do**
    Sample $n_x$ complete factual trajectories using $\pi_{old}$; retain replayable occurrences.
  **end for**
  Retain factual-homogeneous state groups; group eligible occurrences by exact task/observation, choose one physical representative, and retain its original remaining deadline.
  Fit task-excluded Branch prediction model; condition once on current bare labels, scale historical and semantic evidence by learned reliability, and draw aligned task/cluster particles (6), (C.3).
  Solve state-group contrast budget (7)–(9); publish staged factual evidence to memory once.
  **for each allocated occurrence $h$ with $k_h \geq 4$ do**
    Replay the physical prefix; retain the factual child and sample $k_h$ suffixes within $H=L-d_h$.
  **end for**
  Audit physical suffix groups; use only complete allocated groups for likelihood fitting, and commit valid terminal labels once per visited state/depth pool, excluding replayed ancestors.
  *// Physical tree credit and observation-state grouping*
  Back up descendant-leaf terminal means $V_{term}$ (10) and step returns $V_\nu$ (13).
  Build sibling segments; center their scores to obtain $A_{traj}$ (11)–(12).
  Assign deeper groups locally; use factual-root fallback only where no split is assignable.
  Group active turn occurrences by task/observation; compute $A_{step}$ using (14).
  Set all segment coefficients $w_i \leftarrow 1$.
  If the optional plugin is enabled, project $w_i$ by (D.1), conserving parent-group mass.
  Compose $A_i = w_i A_{traj,i} + A_{step,i}$; keep both raw credits and the step coefficient unchanged.
  *// Complete-turn policy update*
  **for each PPO optimizer minibatch do**
    Count global active turns N across ranks; clear accumulated gradients.
    **for each distributed microbatch do**
      Sum token surrogates within turns (15); backpropagate with shared $N$ and reference penalty (D.2), as specified by the total actor loss in Appendix D.2.
    **end for**
    Clip accumulated gradients and take one optimizer step on $\pi_\theta$.
  **end for**
**end for**

## D.1 Optional segment-mass weighting

The core estimator uses unit segment coefficients. An optional segment-balance plugin changes how already-computed trajectory credit is accumulated over unequal-length siblings, without changing the return backup, sibling baseline, or raw advantages. Let $m_{h,c} = \sum_{i \in E(h,c)} v_i$ be a segment's effective turn mass, where $v_i$ is the existing active-sample weight and only positive-mass segments enter an active policy group. When enabled, the plugin targets the sibling mean $\bar{m}_h$ and computes:

$$\bar{m}_h = (1/K)\sum_c m_{h,c}, w_{h,c} = \mathrm{clip}(\lambda_h \bar{m}_h / m_{h,c}, 1/C, C). \quad (D.1)$$

The positive scale $\lambda_h$ is chosen by bounded projection so that $\sum_c m_{h,c} w_{h,c} = \sum_c m_{h,c}$. We use C=6. Without active bounds, $\lambda_h$=1 and every sibling has mass $\bar{m}_h$. Active bounds can leave unequal effective masses while preserving total mass. The coefficient changes only the trajectory term. Appendix B.4 proves existence of the bounded projection, gives a bisection bracket and the zero-mass convention, and confines the guarantee to scalar mass conservation.

The plugin balances positive turn masses, not token counts. It changes only the trajectory coefficient in $A_i^{bal} = w_{h,c} A_{traj}(h,c) + A_{step,i}$ . The step term is unchanged. Equal scalar mass does not imply equal parameter-gradient norms. Appendix B.4 proves the bounded projection's mass conservation.

## D.2 Clipping, reference regularization, and local costs

For token ratio $\rho$ and detached advantage A, $S=\min(\rho A, \mathrm{clip}(\rho, 1-\varepsilon, 1+\varepsilon)A)$. The dual-clipped form is $S_c=S$ for $A \geq 0$ and $S_c=\max(S, cA)$ for $A<0$, with $\varepsilon=0.2$ and c=3. A detached offset centers the displayed PG loss without changing its gradient.

Reference regularization uses a low-variance token penalty with a tangent linear tail. Let $\delta = \log\pi_{ref} - \log\pi_{\theta}$, M=10, and $\delta^* = \log(1+M)$. The penalty is:

$$k_M(\delta) = \begin{cases} \exp(\delta) - \delta - 1 & \text{for } \delta \le \delta^* \\ M - \delta^* + M(\delta - \delta^*) & \text{for } \delta > \delta^* \end{cases}. \quad \text{(D.2)}$$

The minimized actor loss is $-J_{PG} + \beta N^{-1} \sum_{i,j} k_M(\delta_{i,j})$. The tangent tail retains a correcting derivative where a hard cap would saturate. This modified regularizer is distinct from exact reference KL and from the old-policy PPO update monitor. Entropy regularization is disabled in the ALFWorld configuration.

A valid response has exactly one <think>...</think> block followed by one <action>...</action> block, with only whitespace outside them and no characters in the CJK range U+4E00–U+9FFF. The extracted command is lowercased, whitespace-normalized, and checked against current admissible actions; WebShop also accepts a nonempty search query when the page allows search. A formatting violation or inadmissible command incurs local cost one. The extracted command is still submitted to the environment, whose transition determines execution success. Missing action tags trigger the environment wrapper's invalid-command fallback. Costs enter returns before centering; no separate truncation cost is used.

# Appendix E. Evaluation Diagnostics

## E.1 WebShop success and allocation diagnostics

Complete WebShop evaluation yields 81.07% Episode-level success rate (2432/3000). Empirical pass@6 is 87.40% (437/500 goals). These are the 500-goal, six-repeat results of the update-100 checkpoint, separate from the smaller online assessments.

The evaluated WebShop run uses 96 factual roots and 128 added suffixes per update, with likelihood-fitted Root and Branch predictors and local-state-contrast allocation. The historical random-branch run uses the same SFT initialization, task pool, rollout budgets, reward and credit rules, PG aggregation, and reference coefficient 0.05; its branch states are selected randomly with four new suffixes per state. The evaluated lineage covers 156 updates, Figures A3(a) and A3(b) compare the first 100 updates. Figure 4(b) shows its full 156-update history; only the update-100 checkpoint supplies Table A6.

The complete WebShop assessment records 99.88% valid actions (20,079/20,104 turns) and 0.015% output truncation (3 turns). Among 568 failed episodes, 256 reach the 15-action horizon. Failures average 10.41 actions versus 5.84 for successes, locating residual difficulty in task completion as well as interaction execution. The new evaluator stores binary rewards and supplies no native partial-score measure.

Table A6. Complete evaluation of the likelihood-fitted WebShop update-100 checkpoint: 500 goals, six stochastic episodes per goal, and the 1,000-product catalog. The interval resamples whole goal clusters 10,000 times.

| Metric | Result |
|---|---|
| Episode-level success rate (%) | 81.07 (2432/3000) |
| Empirical pass@6 (%) | 87.40 (437/500) |
| Valid actions (%) | 99.88 |
| Horizon termination (%) | 8.67 |
| Mean episode steps | 6.70 |

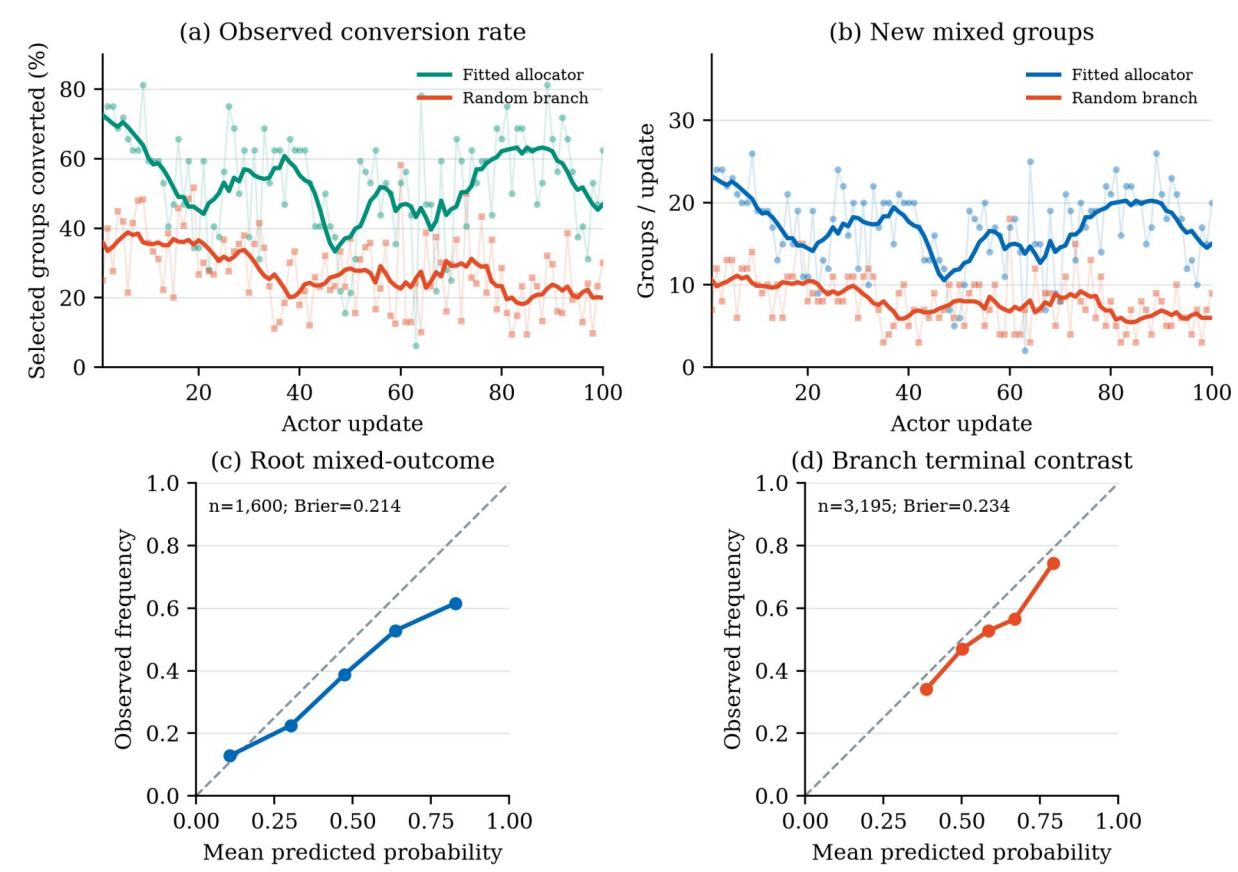

Figure A3. WebShop allocation histories and probability diagnostics. (a) Observed conversion fraction among selected factual-homogeneous observation groups with completed suffix evidence. (b) Newly mixed factual-homogeneous observation groups per update. Both runs use 96 factual roots, 128 new suffixes, and a PPO minibatch size of 240. Thin points are measurements and solid curves are centered nine-record means. (c-d) Root mixed-outcome and Branch terminal-contrast reliability for the fitted run only: five equal-count prediction bins versus subsequent observed group frequency. The dashed diagonal denotes perfect frequency agreement. Root uses 1,600 completed task groups; Branch uses 3,195 selected groups with completed suffix evidence. Brier scores use the same prediction-outcome pairs.

Across updates 1-100, the fitted-allocation run converts an average of 17.00 previously factual-homogeneous observation groups per update, versus 8.03 for random branching; over updates 81-100 the means are 18.50 and 6.00 (Figure A3b). The corresponding per-update conversion fractions among selected factual-homogeneous groups with completed evidence average 53.16% for the fitted run and 27.68% for the random run (Figure A3a); candidate pools differ. Each task/observation key counts once per update. The random run has no completed 3,000-episode assessment, so Table A6 reports only the fitted run.

The stored per-group predictions allow a direct online reliability check (Figure A3c-d). Across the first 100 updates, Root predicts mixed outcomes for 1,600 completed task groups with mean probability 46.94%, versus 37.69% observed (Brier 0.2145). Branch predicts a terminal contrast for 3,195 selected observation groups with completed suffix evidence: 58.66% predicted versus 52.93% observed (Brier 0.2338). Five equal-count prediction bins show overprediction, particularly at high Root scores; these online, selected-group diagnostics do not establish calibration on independent tasks or unselected candidates.

## E.2 Complete evaluation and acquired comparisons

Complete ALFWorld evaluation records 99.17% valid actions and 0.20% truncated turns overall. Unseen Pick is the lowest-success task family at 87.92% (Table 1). Table A7 separates greedy and stochastic outcomes by split; its task-bootstrap intervals resample tasks rather than training seeds.

Table A7. Complete final-model evaluation. The two stochastic success intervals shown are task-cluster bootstrap intervals, not training-seed intervals. Action validity and truncation are turn-level measurements.

| Metric | Decoder | Seen | Unseen |
|---|---|---|---|
| Successful episodes | Greedy | 136/140 | 127/134 |
| Success (%) | Greedy | 97.14 | 94.78 |
| Successful episodes | Stochastic | 1369/1400 | 1295/1340 |
| Success (%) | Stochastic | 97.79 | 96.64 |
| Task-bootstrap 95% interval | Stochastic | 95.93–99.29 | 94.10–98.73 |
| Valid actions (%) | Stochastic | 99.29 | 99.07 |
| Truncated turns (%) | Stochastic | 0.22 | 0.19 |

Selected branch representatives have higher predicted opposite-outcome probabilities than eligible candidates: 47.32% versus 25.93% over 175 ALFWorld updates, and 30.51% versus 18.54% over the first 100 WebShop updates. Among new suffixes, 34.50% and 28.38% end with the label opposite to their factual continuation. These are different statistical units from task success or physical-sibling advantages. For the selected WebShop groups with completed evidence, Root predicts 46.94% versus 37.69% observed mixed outcomes and Branch predicts 58.66% versus 52.93% observed contrasts; prediction is thus overoptimistic in this online diagnostic, especially for high Root scores.

Candidate and selected prediction means cover different representative populations, whereas realized suffix rates cover completed continuations; these pairs do not measure probability calibration. In 13.16% of ALFWorld and 14.43% of WebShop suffixes, the first command and next observation match the factual continuation but the terminal label differs. For WebShop, Figure A3 instead pairs each selected group's advance prediction with its later outcome and reports online reliability diagnostics. Task-level acquisition, suffix outcomes, and the physical sibling advantages in Section 5 remain distinct statistical units.

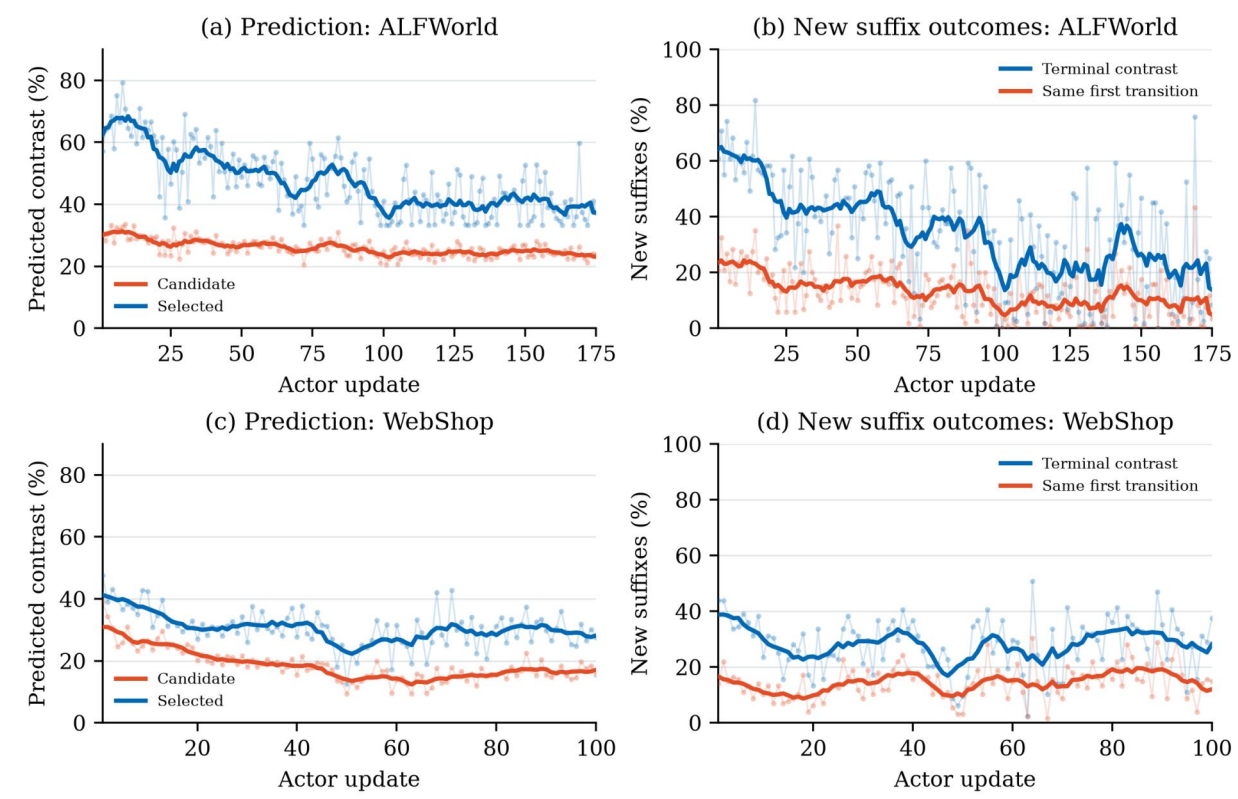

Figure A4. Pre-branch predictions and realized suffix contrasts. Top: the 175-update ALFWorld history; bottom: the evaluated WebShop run through update 100. Left: predicted opposite-outcome probabilities averaged over eligible candidate or selected representatives. Right: percentages of new suffixes with an opposite terminal label, including those whose first command and next observation match the factual continuation. Thin traces and points are measured updates; solid curves are centered nine-record means. Suffix rates divide by the number of completed newly generated suffixes per update.

Table A8. Pre-branch predictions and suffix contrasts. Entries are unweighted means over 175 ALFWorld and 100 WebShop update records. Candidate and selected scores average eligible and awarded task/observation representatives, respectively. Realized rates use completed new suffixes as their denominator; the last row additionally requires the first command and next observation to match the factual continuation.

| Diagnostic | ALFWorld / WebShop |
|---|---|
| Candidate predicted contrast (%) | 25.93 / 18.54 |
| Selected predicted contrast (%) | 47.32 / 30.51 |
| New suffix terminal contrasts (%) | 34.50 / 28.38 |
| Same-first-transition contrasts (%) | 13.16 / 14.43 |

Over the first 100 WebShop updates, the fitted-allocation configuration converts 17.00 previously factual-homogeneous observation groups per update versus 8.03 under historical random branching; selected-group conversion fractions are 53.16% versus 27.68%. Both use the same SFT initialization and rollout budgets. This comparison describes two training configurations (Appendix E.1).

# Appendix F. Observation-Group Density

Over the first 100 actor updates, ALFWorld observation-group density averages 0.424 under the default TSBPO configuration and 0.321 in our GiGPO reference run, a ratio of 1.32 (Table A9; Figure A5). These histories describe the comparison density available to the step baseline; the GiGPO run is separate from the published benchmark values in Table 1.

For update r, let $G_r$ be the task/observation groups produced by the grouping call, including singletons. The numerator $\bar{m}_r$ averages their occurrence counts with equal group weights. The valid-turn denominator $v_r$ is the update's mean generated interaction length $\bar{\ell}_r$ multiplied by its factual valid-action fraction $f_r$. The reported statistic is the per-update ratio $d_r$, averaged equally over R=100 updates:

$$\bar{m}_r = |G_r|^{-1} \sum_{g \in G_r} |g|, v_r = \bar{\ell}_r f_r, \quad d_r = \bar{m}_r / v_r, \bar{d} = R^{-1} \sum_{r=1}^{R} d_r. \tag{F.1}$$

Each included update has one grouping call; Table A9 gives the mean and temporal SD of $d_r$. Repeated visits contribute occurrences and need not be independent trajectories. Our GiGPO reference uses the environment-specific Qwen2.5-1.5B SFT initialization, 16 tasks with eight factual trajectories each (128 roots), no branches, mean-centered GiGPO credit, actor learning rate $3 \times 10^{-6}$, PPO minibatches of 128, and a 50-action horizon. The ratio describes the two histories and is neither an effective sample size nor a gradient-variance estimate.

Table A9. Observation-group density over the first 100 updates of each run. Means and temporal SDs use the per-update definition in Appendix F; singleton groups are included.

| **Diagnostic** | **GiGPO / TSBPO** |
|---|---|
| Mean group size / valid turns | 0.321 / 0.424 |
| Across-record SD | 0.088 / 0.092 |
| Training updates | 100 / 100 |
| Relative to GiGPO mean | 1.00× / 1.32× |

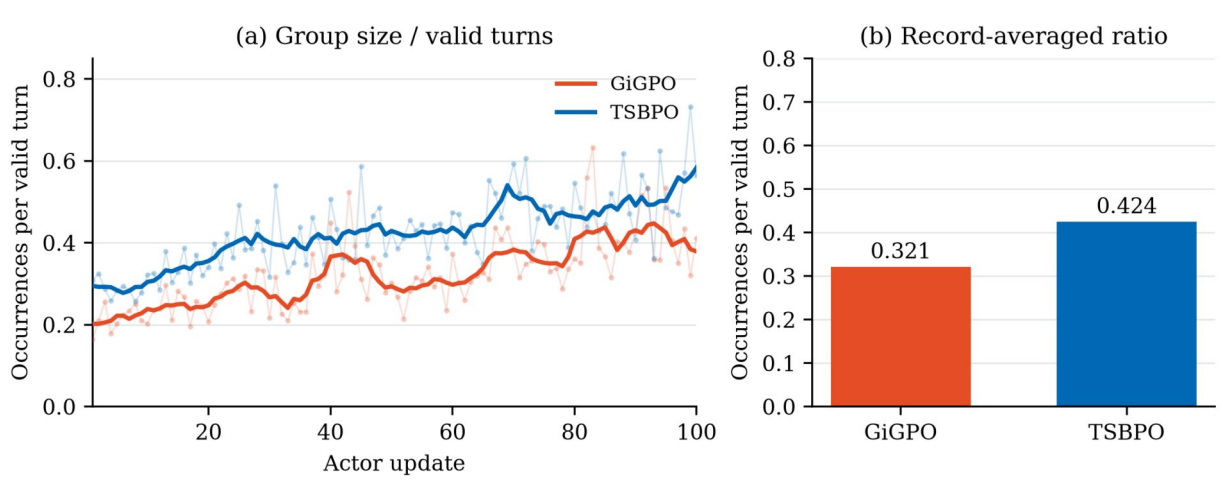

Figure A5. Observation-group density over the first 100 actor updates of each run. Panel (a) shows the per-update ratios defined in Appendix F, with continuous centered seven-record means; panel (b) shows their unweighted means. Each update contains one grouping call. Thin traces and filled markers show measurements. The denominator estimates valid turns per trajectory, rather than the total number of valid turns in a batch.